\def\year{2022}\relax
\documentclass[letterpaper]{article} 
\usepackage{aaai22}  
\usepackage{times}  
\usepackage{helvet}  
\usepackage{courier}  
\usepackage[hyphens]{url}  
\usepackage{graphicx} 
\usepackage{natbib}  
\usepackage{caption} 
\DeclareCaptionStyle{ruled}{labelfont=normalfont,labelsep=colon,strut=off} 
\usepackage{algorithm}
\usepackage{algorithmicx}

\usepackage[noend]{algpseudocode}
\algnewcommand{\IfThenElse}[3]{
  \State \algorithmicif\ #1\ \algorithmicthen\ #2\ \algorithmicelse\ #3}
  
\usepackage{enumitem}
\usepackage{balance}
\usepackage{amsmath}

\usepackage{graphicx}
\graphicspath{{img/}}
\usepackage{caption}
\usepackage[export]{adjustbox}

\usepackage{lscape}

\usepackage{xcolor}
\usepackage{tikz}
\usetikzlibrary[
                positioning,
                fit,
                arrows.meta,
                calc,
                patterns,
                shapes.arrows,
                   ]
\usepackage{amsmath}
\usetikzlibrary{math, patterns}
\usepackage{booktabs}

\usepackage{newfloat}
\usepackage{listings}
\floatstyle{ruled}
\newfloat{listing}{tb}{lst}{}
\floatname{listing}{Listing}
\title{S-Cyc: A Learning Rate Schedule for Iterative Pruning of ReLU-based Networks}
\author {
    Shiyu Liu,
    Chong Min John Tan,
    Mehul Motani
}
\affiliations {
    Department of Electrical and Computer Engineering \\
    School of Engineering, National University of Singapore \\
    shiyu\_liu@u.nus.edu, e0441892@u.nus.edu, motani@nus.edu.sg
}

\usepackage{bibentry}

\makeatletter
\newcommand*{\rom}[1]{\expandafter\@slowromancap\romannumeral #1@}
\makeatother

\begin{document}

\setlength{\belowdisplayskip}{0pt} \setlength{\belowdisplayshortskip}{0pt}
\setlength{\abovedisplayskip}{0pt} \setlength{\abovedisplayshortskip}{0pt}

\maketitle

\begin{abstract}
We explore a new perspective on adapting the learning rate (LR) schedule to improve the performance of the ReLU-based network as it is iteratively pruned. Our work and contribution consist of four parts: (i) We find that, as the ReLU-based network is iteratively pruned, the distribution of weight gradients tends to become narrower. This leads to the finding that as the network becomes more sparse, a larger value of LR should be used to train the pruned network. (ii) Motivated by this finding, we propose a novel LR schedule, called S-Cyclical (S-Cyc) which adapts the conventional cyclical LR schedule by gradually increasing the LR upper bound ($\texttt{max\_lr}$) in an $\texttt{S}$-shape as the network is iteratively pruned.
We highlight that S-Cyc is a method agnostic LR schedule that applies to many iterative pruning methods. (iii) We evaluate the performance of the proposed S-Cyc and compare it to four LR schedule benchmarks. Our experimental results on three state-of-the-art networks (e.g., VGG-19, ResNet-20, ResNet-50) and two popular datasets (e.g., CIFAR-10, ImageNet-200) demonstrate that S-Cyc consistently outperforms the best performing benchmark with an improvement of 2.1\% - 3.4\%, without substantial increase in complexity. (iv) We evaluate S-Cyc against an oracle and show that S-Cyc achieves comparable performance to the oracle, which carefully tunes $\texttt{max\_lr}$ via grid search.

\end{abstract}

\section{Introduction}
Network pruning is a process of removing weights, filters or neurons from neural networks \cite{lecun1990optimal,han2015learning,li2016pruning}. Several state-of-the-art pruning methods \cite{renda2020comparing,frankle2018lottery,frankle2020linear} have demonstrated that a large amount of parameters can be removed without sacrificing accuracy. This greatly reduces the resource demand of neural networks, such as storage requirements and energy consumption \cite{han2015learning,li2016pruning}. 

The inspiring performance of pruning methods hinges on a key factor - Learning Rate (LR) - as mentioned in prior works \cite{renda2020comparing,frankle2018lottery}. Specifically, \citeauthor{frankle2018lottery} (2019) propose the concept of the Lottery Ticket Hypothesis and demonstrate that the winning tickets (i.e., the pruned subnetwork that can train in isolation to full accuracy) cannot be found without applying a LR warmup schedule. In a follow-up work, \citeauthor{renda2020comparing} (2020) propose LR rewinding which rewinds the LR schedule to its initial state during iterative pruning and demonstrate that it can outperform standard fine-tuning. Overall, these two works suggest that LR plays an important role in network pruning. 

In this paper, we take the investigation one step further and explore a new perspective on adapting the LR schedule to improve the iterative pruning performance of the ReLU-based network. Our contributions are as follows.

\begin{enumerate}[noitemsep,leftmargin=5mm, topsep=-4pt]
\item Via extensive experiments on various networks, datasets and pruning methods, we find that, as the ReLU-based network is iteratively pruned, the distribution of weight gradients tends to become narrower, suggesting that a larger value of LR should be used for pruning.
\item We propose a novel LR schedule called the S-Cyclical (S-Cyc) which adapts the conventional cyclical LR \cite{smith2017cyclical} by gradually increasing the LR upper bound ($\texttt{max\_lr}$) in an \texttt{S}-shape as the ReLU-based neural network is iteratively pruned.

\item We compare the performance of S-Cyc to four LR schedule benchmarks on three state-of-the-art networks (e.g., VGG-19 \cite{vgg16}, ResNet-20, ResNet-50 \cite{resnet18}) and two popular datasets (e.g., CIFAR-10 \cite{cifar10}, ImageNet-200 \cite{tinyimagenet}). Our experimental results demonstrate S-Cyc consistently outperforms the best performing benchmark  with an improvement of 2.1\% - 3.4\%.

\item We investigate the trajectory of $\texttt{max\_lr}$ at each pruning cycle. Our results demonstrate that the value of $\texttt{max\_lr}$ estimated by S-Cyc is very competitive, i.e., it can achieve comparable performance to a greedy oracle which carefully tunes $\texttt{max\_lr}$ via grid search.

\end{enumerate}

\section{Network Pruning and Learning Rate}
In Section \ref{sec2.1}, we first review prior work on network pruning. Next, in Section \ref{sec2.2}, we highlight the important role of LR in network pruning and position our work accordingly.

\subsection{Prior Works on Network Pruning}
\label{sec2.1}

Network pruning is an established idea dating back to 1990 \cite{lecun1990optimal}. The motivation is that neural networks tend to be overparameterized and redundant weights can be removed with a negligible loss in accuracy
\cite{arora2018optimization,allenzhu2018convergence,denil2013predicting}. Given a trained network, one {\bf pruning cycle} consists of three steps as follows.
\begin{enumerate}[noitemsep,leftmargin=5mm, topsep=0pt]
\item Prune the network according to certain heuristics.
\item Freeze pruned parameters as zero.
\item Retrain the pruned network to recover the accuracy.
\end{enumerate}
Repeating the pruning cycle multiple times until the target sparsity or accuracy is met is known as {\bf iterative pruning}. Doing so often results in better performance than one-shot pruning (i.e., perform only one pruning cycle) \cite{han2015learning,frankle2018lottery, li2016pruning}. There are two types of network pruning - unstructured and structured - which will be discussed in detail below.

\noindent {\bf Unstructured Pruning: } Unstructured pruning removes individual weights according to certain heuristics such as magnitude \cite{han2015learning,frankle2018lottery} or  gradient \cite{hassibi1993second, lecun1990optimal, lee2018snip, xiao2019autoprune, theis2018faster}. Examples are \cite{lecun1990optimal}, which performs pruning based on the Hessian Matrix, and \cite{theis2018faster}, which uses Fisher information to approximate the Hessian Matrix.
Similarly, \citeauthor{han2015learning} (2015) remove weights with the smallest magnitude and this approach was further incorporated with the three-stage iterative pruning pipeline in \cite{han2015deep}.

\noindent {\bf Structured Pruning: } Structured pruning involves pruning weights in groups, neurons, channels or filters \cite{He2019median, molchanov2016pruning, molchanov2019importance, luo2017thinet, yu2018nisp, tan2020dropnet,wang2020dynamic,lin2020hrank}. Examples are \cite{hu2016network}, which removes neurons with high average zero output ratio, and \cite{li2016pruning}, which prunes neurons with the lowest absolute summation values of incoming weights. More recently, \citeauthor{yu2018nisp} (2018) propose the neuron importance score propagation algorithm to evaluate the importance of network structures. \citeauthor{molchanov2019importance} (2019) use Taylor expansions to approximate a filter's contribution to final loss and \citeauthor{Wang_2020_CVPR} (2020) optimize the neural network architecture, pruning policy, and quantization policy in a joint manner.

\noindent {\bf Other Works: } In addition to works mentioned above, several other works also share some deeper insights in network pruning \cite{liu2018rethinking, zhu2017prune, liu2018darts,wang2020pruning}. For example, \citeauthor{liu2018darts} (2018) demonstrate that training-from-scratch on the right sparse architecture yields better results than pruning from pre-trained models. Similarly, \citeauthor{wang2020pruning} (2020) suggest that the fully-trained network could reduce the search space for the pruned structure. More recently, \citeauthor{yeom2021pruning} (2021) evaluate the importance of the network structure using their relevance computed via the concepts of explainable AI, connecting interpretability to model compression. \citeauthor{luo2020neural} (2020) address the issue of pruning residual connections with limited data and \citeauthor{ye2020good} (2020) theoretically prove the existence of small subnetworks with lower loss than the unpruned network. One milestone paper \cite{frankle2018lottery} points out that reinitializing with the original parameters plays an important role in network pruning and helps to further prune the network with negligible loss in accuracy. Some follow-on works \cite{zhou2019deconstructing,franklestabilizing,renda2020comparing,malach2020proving} investigate this phenomenon more precisely and apply this method in other fields (e.g., transfer learning \cite{mehta2019sparse}, reinforcement learning and natural language processing \cite{yu2019playing}). 

\subsection{The Important Role of LR and Our Work}
\label{sec2.2}
Several recent works \cite{renda2020comparing,frankle2018lottery} have noticed the important role of LR in network pruning. For example, \citeauthor{frankle2018lottery} (2019) demonstrate that training VGG-19 \cite{vgg16} with a LR warmup schedule (i.e., increase LR to 1$\texttt{e}$-1 first and decrease it to 1$\texttt{e}$-3) and a constant LR of 1$\texttt{e}$-2 results in comparable accuracy for the unpruned network. However, as the network is iteratively pruned, the LR warmup schedule leads to a much higher accuracy (see Fig.7 in \cite{frankle2018lottery}).   
One follow-up work \cite{renda2020comparing} further investigates this phenomenon and proposes a retraining technique called LR rewinding. They demonstrate that LR rewinding can always outperform the standard retraining technique called fine-tuning \cite{han2015learning}. The difference is that fine-tuning trains the unpruned network with a LR warmup schedule, and then it retrains the pruned network with a constant LR (i.e., the final LR of the LR warmup schedule) in subsequent pruning cycles \cite{liu2018rethinking}. LR rewinding retrains the pruned network by rewinding the LR warmup schedule to its initial state, which is equivalent to using the same LR warmup schedule for subsequent pruning cycles. As an example, they train the ResNet-50 \cite{resnet18} with a LR warmup schedule (i.e., increase LR to 0.4 and decrease it to 4\texttt{e}-4). The performance of LR rewinding and fine-tuning are comparable for the unpruned network. However, as the network is iteratively pruned, LR rewinding leads to a much higher accuracy (see Figs.1 \& 2 in \cite{renda2020comparing}). Overall, both of these two works suggest that LR plays an important role in network pruning. 

\begin{figure*}[!t]
\hspace{4mm}\begin{minipage}{0.45\textwidth}
\includegraphics[width=0.9\linewidth, left]{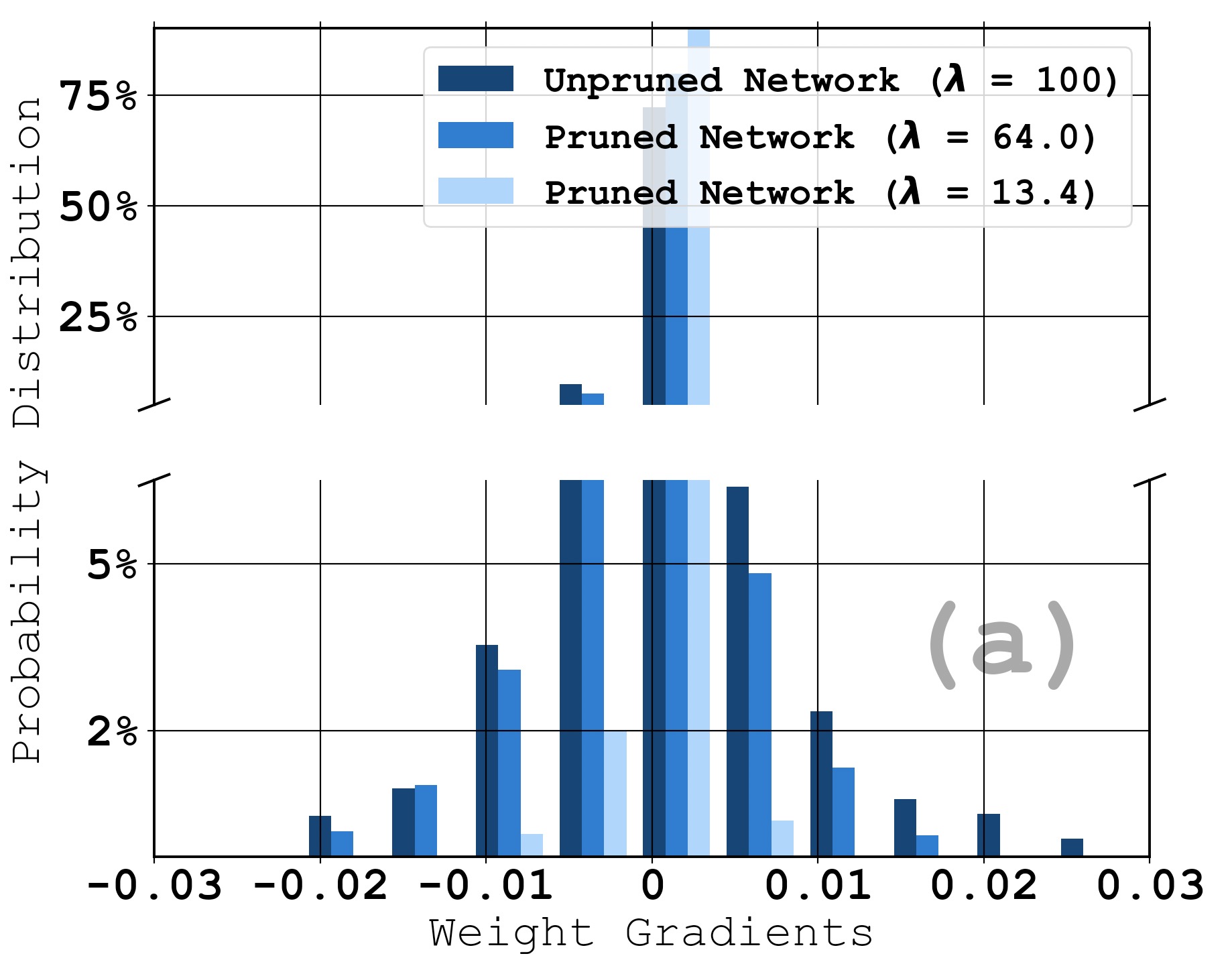}
\end{minipage}%
\hspace{4mm}\begin{minipage}{0.45\textwidth}
\includegraphics[width=0.9\linewidth, right]{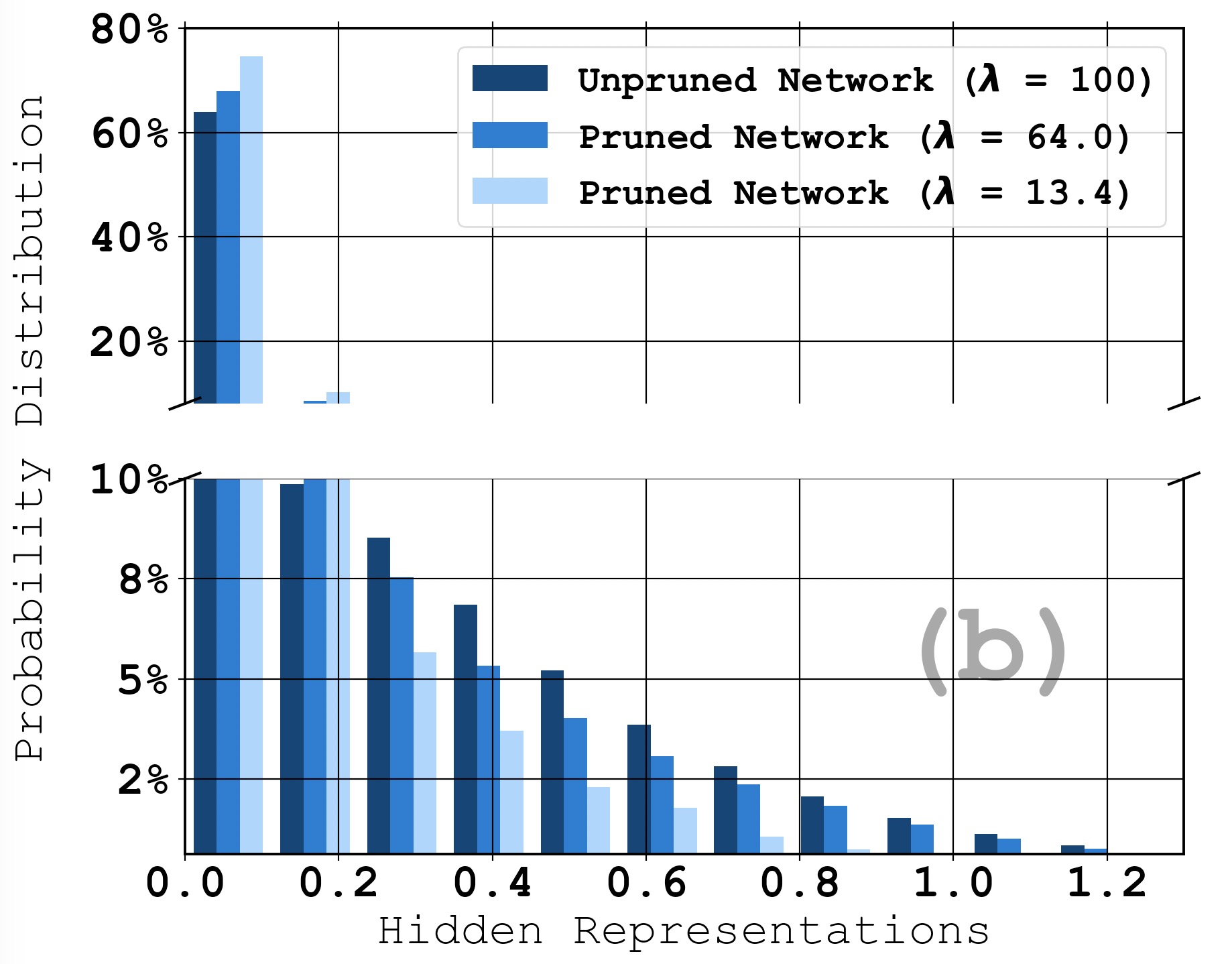}
\end{minipage}
\vspace{-2mm}
\caption{ (a) The distribution of all weight gradients when iteratively pruning a fully connected ReLU-based network using global magnitude pruning method \cite{han2015learning}, where $\lambda$ is the percent of weights remaining. (b) The corresponding distribution of hidden representations (i.e., post-activation outputs of all hidden layers). Please note that there is a line breaker in the vertical axis. We note that this finding generalizes to other networks, datasets and pruning methods (see Section \ref{3.3}).} 
\label{mlp_no_norm}
\vspace{-2mm}
\end{figure*}

{\bf Our work }takes the investigation one step further and explores a new perspective on adapting the LR schedule to improve the iterative pruning performance of the ReLU-based network. The proposed LR schedule is method agnostic and applies to many iterative pruning methods. We chose to focus on iterative pruning of ReLU-based networks for two reasons: (i) Iterative pruning tends to provide better pruning performance than one-shot pruning as reported in the literature \cite{frankle2018lottery, renda2020comparing}. (ii) ReLU has been widely used in many state-of-the-art neural networks (e.g., ResNet, VGG). Those networks have achieved outstanding performance in various tasks \cite{resnet18, vgg16}.

\section{A New Insight on Network Pruning}
\label{sec3}
In this section, we first provide a new insight in network pruning using empirical results in Section \ref{3.1}. In Section \ref{3.2}, we provide a possible explanation for the new insight. In Section \ref{3.3}, we highlight that this new insight can be generalized to other networks, datasets and pruning methods. 

\subsection{An Experiment and a New Insight}
\label{3.1}

We first design a fully connected ReLU-based neural network with three hidden layers of 256 neurons each. We train the neural network using a randomly selected subset (80\%) from the Fashion-MNIST dataset via SGD \cite{ruder2016overview} (LR = 1\texttt{e}-3, momentum = 0.9) with a batch size of 64 for 500 epochs. We note that all weights are initialized using He Initialization \cite{he2015delving} and all hyper-parameters are tuned for performance via grid search (e.g., LR from 1\texttt{e}-4 to 1\texttt{e}-2). 
We apply the global magnitude pruning \cite{han2015learning} (i.e., removing weights with the smallest magnitudes anywhere in the network) with a pruning rate of 0.2 (i.e., prune 20\% of the remaining parameters) to iteratively prune the network for 10 pruning cycles and plot the distribution of all weight gradients at the end of training in Fig. \ref{mlp_no_norm} (a), where $\lambda$ is the percent of weights remaining.
There are 10 visible bins estimated by the the Sturges Rule and the edge values rang from -0.021 to 0.027 with a bin width of 0.004.

In Fig.\ref{mlp_no_norm} (a), we observe that the distribution of weight gradients tends to become less heavy-tailed as the network is iteratively pruned. For example, the unpruned network ($\lambda$ = 100) has more than 6\% of weight gradients with values greater than 0.009 (rightmost 4 bars) or less than -0.011 (leftmost 2 bars), while the pruned network ($\lambda = 13.4$) has less than 1\% of weight gradients falling into those regions.

During the training of network, the weight update of $w_i$ is
\begin{equation}
w_i \leftarrow w_i +  \alpha \frac{\partial \mathcal{L}}{\partial w_i},
\end{equation}
where $\alpha$ is the LR and $\mathcal{L}$ is the loss function. 

\noindent {\bf New Insight:}  Assume that $\alpha$ is well-tuned to ensure the weight update (i.e.,  $\alpha \frac{\partial \mathcal{L}}{\partial w_i}$)  is sufficiently large to prevent the network from getting stuck in local optimal points \cite{bengio2012practical, goodfellow2016deep}. In Fig. \ref{mlp_no_norm} (a), the magnitude of weight gradients (i.e., $\frac{\partial \mathcal{L}}{\partial w_i}$) tends to decrease as the network is iteratively pruned. To preserve the same weight updating size and effect as before, {\em a larger value of LR ($\alpha$) should be used to train the pruned network.}

\subsection{Result Analysis}
\label{3.2}
We now provide a possible explanation for the change in the distribution of weight gradients. We assume each $x_i w_i$ (i.e., the scalar neuron input $\times$ its associated scalar weight) is an i.i.d. random variable. Then, the variance of the neuron's pre-activation output ($ \sum_{i=1}^{n} x_i w_i$, $n$ is the number of inputs) will be $\sum_{i=1}^{n} \text{Var} (x_i w_i)$. Pruning the network is equivalent to reducing the number of inputs from $n$ to $n - k$, resulting in a smaller variance of $\sum_{i=1}^{n-k} \text{Var} (x_i w_i)$. Hence, the distribution of the pre-activation output after pruning is narrower. Since ReLU returns its raw input if the input is positive, the distribution of hidden representations (i.e., post-activation outputs) becomes narrower as well. This can be verified from Fig. \ref{mlp_no_norm} (b), where we plot the distribution of hidden representation (i.e., post-activation output of all hidden layers) from the experiment in Section \ref{3.1}. We note that the weight gradient $\frac{\partial \mathcal{L}}{\partial w_i} $ is proportional to the hidden representation $x_i$ that associates with $w_i$ (i.e., $\frac{\partial \mathcal{L}}{\partial w_i} \propto x_i$). Therefore, as the network is iteratively pruned, the distribution of hidden representations tends to become narrower (see Fig. \ref{mlp_no_norm} (b)), leading to a narrower distribution of weight gradients (see Fig. \ref{mlp_no_norm} (a)). As a result, a larger value of LR should be used to ensure the weight update is sufficiently large.

\begin{figure*}[!t]
\begin{minipage}{0.48\linewidth}
\centering
\resizebox{1.0\linewidth}{!}{
\begin{tikzpicture}[yscale=2.5,xscale=0.4]
\draw[line width = 0.5mm](-6,-0.05)--(-0.6,-0.05);
\draw[line width = 0.5mm](-0.3,-0.05)--(10.4,-0.05);
\draw[line width = 0.5mm](-0.55,-0.13)--(-0.55,0.04);
\draw[line width = 0.5mm](-0.3,-0.13)--(-0.3,0.04);
\draw[line width = 0.5mm](10.6,-0.13)--(10.6,0.03);
\draw[line width = 0.5mm](10.4,-0.13)--(10.4,0.03);
\draw[line width = 0.4mm,->](10.6,-0.05)--(22.5,-0.05);
\draw[line width = 0.4mm,->](-6,-0.05)--(-6,2.35)node[right=0.4cm,below=0.1cm]{};
\draw[line width = 0.4mm,->](-6,2.3)--(22.5,2.3)node[right=-3cm, below=0.1cm]{};
\node[rotate=90,above=0.1cm]at (-6,1.0){\Large Learning Rate};
\draw[domain=1.5:18, samples=100,thick, red,line width=2pt] plot (\x,{(1/(1+exp((-\x+8)))+0.7});
\node[above=0.1cm]at (8,1.95){\Large $\lambda$ = 100 $\leftarrow$ Percent of Weights Remaining $\rightarrow$ $\lambda$ = 0};


\draw[red,line width=2pt](-4.5,0.7)--(1.5,0.7);
\draw[blue, line width = 0.5mm](-6,-0.05)--(-4.5,0.71)
--(-4,0.71)
--(-4,0.071)
--(-2.5,0.071)
--(-2.5,0.0072)
--(-1,0.0072)
--(-1,0)
--(-1,-0.04)
;

\draw[blue, line width = 0.5mm](0,-0.05)--(1.5,{(1/(1+exp((6.5)))+0.7})
--(2,{(1/(1+exp((6.5)))+0.7})
--(2,{((1/10)*(1/(1+exp((6.5)))+0.7})
--(3.5,{((1/10)*(1/(1+exp((6.5)))+0.7})
--(3.5,{((1/100)*(1/(1+exp((6.5)))+0.7})
--(5,{((1/100)*(1/(1+exp((6.5)))+0.7})
--(5,{((1/1000)*(1/(1+exp((6.5)))+0.7})
--(5,-0.04)
;

\draw[blue, line width = 0.5mm](5,-0.05)--(6.5,{(1/(1+exp((1.5)))+0.7})
--(7,{(1/(1+exp((1.5)))+0.7})
--(7,{((1/10)*(1/(1+exp((1.5)))+0.7})
--(8.5,{((1/10)*(1/(1+exp((1.5)))+0.7})
--(8.5,{((1/100)*(1/(1+exp((1.5)))+0.7})
--(10,{((1/100)*(1/(1+exp((1.5)))+0.7})
--(10,{((1/1000)*(1/(1+exp((1.5)))+0.7})
--(10,-0.04)
;

\draw[blue, line width = 0.5mm](11.5,-0.05)--(13,{(1/(1+exp((-5)))+0.7})
--(13.5,{(1/(1+exp((-5)))+0.7})
--(13.5,{((1/10)*(1/(1+exp((-5)))+0.7})
--(15,{((1/10)*(1/(1+exp((-5)))+0.7})
--(15,{((1/100)*(1/(1+exp((-5)))+0.7})
--(16.5,{((1/100)*(1/(1+exp((-5)))+0.7})
--(16.5,{((1/1000)*(1/(1+exp((-5)))+0.7})
--(16.5,-0.04)
;
\draw[blue, line width = 0.5mm](16.5,-0.05)--(18,{(1/(1+exp((-10)))+0.7})
--(18.5,{(1/(1+exp((-10)))+0.7})
--(18.5,{((1/10)*(1/(1+exp((-10)))+0.7})
--(20,{((1/10)*(1/(1+exp((-10)))+0.7})
--(20,{((1/100)*(1/(1+exp((-10)))+0.7})
--(21.5,{((1/100)*(1/(1+exp((-10)))+0.7})
--(21.5,{((1/1000)*(1/(1+exp((-10)))+0.7})
--(21.5,-0.04)
;
%
\draw[line width = 0.4mm,<->](-6,-0.13)--(-1,-0.13);
\draw[line width = 0.4mm,<->](0,-0.13)--(5,-0.13);
\draw[line width = 0.4mm,<->](5,-0.13)--(10.05,-0.13);
\draw[line width = 0.4mm,<->](11.5,-0.13)--(16.5,-0.13);
\draw[line width = 0.4mm,<->](16.5,-0.13)--(21.55,-0.13);

\foreach \P/\N in {(-3.4,-0.13)/{\small Pruning Cycle} $\texttt{0}$,(2.5,-0.13)/$\texttt{q}$,
(7.9,-0.13)/$\texttt{q} + 1$,(14.3,-0.13)/$\texttt{L}-2$,
(19.3,-0.13)/$\texttt{L}-1$}{
\node[below,align=left] at \P {\Large \N};
}

\foreach \M in {(-4.5, 3.5), (1.5,3.5),(6.5,4.4),(13,8.45),(18,8.5)}{
\filldraw[yscale=0.2] \M circle (3pt);
}

\foreach \M in {(-0.9,1.5),(-.5,1.5),(-0.1,1.5),(10,1.5),(10.4,1.5),(10.8,1.5)}{
\filldraw[yscale=0.20] \M circle (2pt);
}

\draw[line width = 0.5mm, dashed](-4.5,0.69)--(20,0.69);
\draw[line width = 0.5mm, dashed](18,1.7)--(20,1.7);

\node[above=0.1cm]at (20.9, 1.55){\Large  $\delta$ + $\epsilon$};
\node[above=0.1cm]at (20.3, 0.57){\Large  $\epsilon$};

\node[above=0.1cm]at (-2.2, 0.45){\large $\texttt{max\_lr}$};
\node[above=0.1cm]at (3.8, 0.45){\large $\texttt{max\_lr}$};
\node[above=0.1cm]at (8.9, 0.65){\large $\texttt{max\_lr}$};
\node[above=0.1cm]at (15.4, 1.4){\large $\texttt{max\_lr}$};
\node[above=0.1cm]at (20.4, 1.4){\large $\texttt{max\_lr}$};

\draw[->,>=latex,thick,xscale=1.5] (-3.5,0.8) -- (-1,0.8);
\node[above,rotate=0] at (-3.3,0.8){\texttt{No Growth}};

\draw[->,>=latex,thick,xscale=1.5] (1.5,0.75) -- (4,0.9);
\node[above,rotate=15,align=left] at (4,0.82){\texttt{Slow}\\\texttt{Growth \rom{1}}};

\draw[->,>=latex,thick] (7,1.1) -- (8.5,1.5);
\node[above,rotate=60,align=left]at (7.7,1.3){\texttt{Fast}\\ \texttt{Growth}};
\draw[->,>=latex,thick] (12,1.75) -- (16.5,1.75); \node[above,rotate=0]at (14.6,1.74){\texttt{Slow Growth \rom{2}}};

\end{tikzpicture}}
\end{minipage}%
\hspace*{6mm}\begin{minipage}{.48\linewidth}
\centering
\setlength{\textfloatsep}{0pt}
\vspace*{-5mm} \begin{algorithm}[H]
{\fontsize{9.5pt}{9.5pt}\selectfont
    \caption{Algorithm of the Proposed S-Cyc} 
    \label{algorithm1}
    \begin{algorithmic}[1]
    	\Require (1) lower bound $\epsilon$, (2) upper bound $\delta + \epsilon$,  (3) pruning rate \texttt{p}, (4) number of pruning cycles $\texttt{L}$, (5) number of training epochs $\texttt{t}$, (6) S-shape control term $\beta$, (7) delay term $\texttt{q}$.
    	\For {$ \texttt{m} = 0$ to $\texttt{L}$}
    		\If {$\texttt{m} \leq \texttt{q}$} $\texttt{max\_lr} = \epsilon $
   	 	\Else {} $\texttt{max\_lr} = \frac{\delta}{1 + (\frac{\gamma}{1-\gamma})^{-\beta}} + \epsilon$, where $\gamma = 1-(1-\texttt{p})^{m - \texttt{q}}$
   	 	\EndIf
		\For {$ \texttt{i} = 0$ to $\texttt{t}$}
			\State linearly warmup learning rate to $\texttt{max\_lr}$
			\State then drop it multiple times at certain epochs
		\EndFor
    	\EndFor
    \end{algorithmic}}
\end{algorithm}
\end{minipage}
\vspace{-0mm}
\caption{Illustration of the Proposed S-Cyc Learning Rate Schedule (left) and its detailed algorithm (right).}
\vspace{-0mm}
\label{SC}
\end{figure*}

\subsection{More Experimental Results}
\label{3.3}

{\bf Effect of Batch Normalization: } Batch Normalization (BN) \cite{ioffe2015batch} is a popular technique to reformat the distribution of hidden representations, so as to address the issue of internal covariate shift. To examine the effect of BN , we apply BN ($\mu$ = 0, $\sigma^2$ = 1) to each hidden layer (before ReLU) and re-plot the distribution of hidden representations and weight gradients (see Fig. \ref{mlp_distribution_norm} in the \textbf{Appendix}).
We note that BN slightly reduces the change in the distribution of hidden representations and weight gradients as the network is iteratively pruned, but the trend still largely mirrors those in Fig. \ref{mlp_no_norm} (a) \& (b). As an example,  the unpruned network ($\lambda$ = 100) has about 3\% of weight gradients with values greater than 0.03 (rightmost bar) or less than -0.024 (leftmost bar), while the pruned network ($\lambda$ = 13.4) has less than 1\% of weight gradients falling into those regions. 

\noindent {\bf Other Pruning Methods, Datasets and Networks: } In addition to the global magnitude pruning, two unstructured pruning methods (i.e., layer magnitude, global gradient) suggested by \cite{Blalock2020} and one structured pruning method (\texttt{L1} norm pruning) \cite{li2016pruning} are examined as well. Those methods are used to iteratively prune AlexNet \cite{krizhevsky2010convolutional}, ResNet-20 and VGG-16 on the CIFAR-10 dataset. The network parameters are summarized in Table. \ref{arch_sec2} in the \textbf{Appendix}. The results shown in Figs. \ref{alex} - \ref{vgg} in the \textbf{Appendix} largely mirror those in Figs. \ref{mlp_no_norm} as well.

\section{A New Learning Rate Schedule}
\label{sec4}
In Section \ref{sec4.1}, we review existing works on LR and shortlist four benchmarks for comparison. In Section \ref{sec4.2}, we introduce the idea of S-Cyc and highlight the difference with existing works. 
In Section \ref{sec4.3}, we detail the S-Cyc algorithm.

\subsection{LR Schedule Benchmarks}
\label{sec4.1}

Learning rate is the most important hyper-parameter in training networks \cite{ goodfellow2016deep}. The LR schedule is to adjust LR during training by a pre-defined schedule. Common LR schedules are
\begin{enumerate}[noitemsep,leftmargin=5mm, topsep=0pt]
\item LR Decay starts with a large initial LR and decays it by a certain factor after a pre-defined number of epochs. Several works \cite{you2019does,ge2019step,an2017exponential} have demonstrated that decaying LR helps the network converge better and avoids undesired oscillations.
\item LR Warmup is to increase the LR to a large value over certain epochs and then decreases the LR by a certain factor. It is a popular schedule used by many practitioners for transfer learning \cite{he2019bag} and network pruning \cite{frankle2018lottery,frankle2020linear}.
\item Cyclical LR \cite{smith2017cyclical} is to vary the LR cyclically between a pre-defined lower bound and upper bound. LR Warmup can be considered as a special form of the Cyclical LR. The difference is that the LR Warmup does only one up-and-down cycle while the Cyclical LR repeats the up-and-down cycle multiple times.
\end{enumerate}

All of the three LR schedules and constant LR will be used as {\bf benchmarks for comparison}. For the proposed S-Cyc, we evaluate its performance using SGD with momentum = 0.9 and a weight decay of 1$\texttt{e}$-4 (same as \cite{renda2020comparing,frankle2018lottery}). 

We note that, in addition to LR schedules which vary LR by a pre-defined schedule, adaptive learning rate optimizers such as AdaDelta \cite{zeiler2012adadelta}, Adam \cite{kingma2014adam} and RMSProp \cite{tieleman2012lecture}  provide heuristic based approaches to adaptively vary the step size of weight update based on observed statistics of the past gradients. All of them are sophisticated optimization algorithms and much work \cite{gandikota2021vqsgd,jentzen2021strong} has been done to investigate their behaviours and mechanisms. The effect of those adaptive LR optimizers on the proposed S-Cyc will be discussed in Section \ref{discussion}.

\subsection{S-Cyclical Learning Rate Schedule}
\label{sec4.2}
In Sections \ref{3.2}, we find that as the ReLU-based network is iteratively pruned, the distribution of weight gradients becomes narrower, and a larger value of LR should be used. This motivates us to propose S-Cyc LR schedule for iterative pruning of ReLU-based networks. As illustrated in Fig. \ref{SC}, the main idea of the proposed S-Cyc is to apply the LR warmup schedule for every pruning cycle, with a gradual increase of the LR upper bound (i.e., \texttt{max\_lr}) in an \texttt{S}-shape as the network is iteratively pruned. The LR warmup schedule for one pruning cycle first increases the value of LR to \texttt{max\_lr}, maintains its value, then decreases it sharply. This LR warmup schedule is meant to be flexible and can change depending on different neural networks and datasets.

\noindent The proposed S-Cyc consists of four phases as follows:
\begin{enumerate}[noitemsep,leftmargin=5mm, topsep=0pt]
\item In the phase of {\bf No Growth}, we do not increase \texttt{max\_{lr}} until the pruning cycle \texttt{q} (see Fig. \ref{SC}), where \texttt{q} is a tunable parameter. It is because the unpruned network usually contains a certain amount of weights with zero magnitude. Those parameters are likely to be pruned at the first few pruning cycles, and removing such weights has negligible effect on the distribution of weight gradients. For example, in Fig. \ref{alex} in the \textbf{Appendix}, we compare the pruned network ($\lambda$ = 80) to the unpruned network and observe very little difference in the distribution.

\item In the phase of {\bf Slow Growth \rom{1}}, we have removed most zero magnitude weights and started pruning weights with small magnitude. Pruning such weights has a small effect on the distribution of weight gradients. Hence, we slightly increase \texttt{max\_{lr}} after pruning cycle \texttt{q}.  

\item In the phase of {\bf Fast Growth}, we greatly increase \texttt{max\_{lr}}. It is because that we start removing weights with large magnitude and the distribution of weight gradients becomes much narrower. Hence, the LR needs to be increased to ensure the weight update is meaningful. 

\item In the phase of {\bf Slow Growth \rom{2}}, the network is heavily pruned and very few parameters left in the neural network. By using the same pruning rate, a very small portion of the weights will be pruned. This could cause a marginal effect on the distribution of weight gradients. Hence, we only need to slightly increase \texttt{max\_{lr}}.

\end{enumerate}

We note that the design of S-Cyc is based on the assumption that existing pruning methods tend to prune weights with small magnitude. The proposed S-Cyc is modified from the conventional cyclical LR schedule \cite{smith2017cyclical}. 
{\bf The key difference with existing LR schedules} (e.g., cyclical LR, LR warmup) is that S-Cyc is adaptive and increases the value of \texttt{max\_{lr}} as the network is iteratively pruned, while existing LR schedules do not factor in the need to change \texttt{max\_{lr}} during different pruning cycles.

\subsection{Implementation of S-Cyc LR Schedule}
\label{sec4.3}
As for the implementation of the proposed S-Cyc LR schedule, we designed a function as shown below.
\begin{equation}
\texttt{max\_lr} = \frac{\delta}{1 + (\frac{\gamma}{1-\gamma})^{-\beta}} + \epsilon, 
\end{equation}
where $\gamma = 1 - (1 - \texttt{p})^{m - \texttt{q}}$ is the input of the function and $\texttt{max\_lr}$ is the output of the function. The parameter \texttt{p} is the pruning rate and $\texttt{m}$ is the number of completed pruning cycles. The parameters $\beta$ and $\texttt{q}$ are used to control the shape of the \texttt{S} curve. The larger the $\beta$, the later the curve enters the \texttt{Fast Growth} phase. The parameter \texttt{q} determines at which pruning cycle the S-Cyc enters the \texttt{Slow Growth \rom{1}} phase. When $q = 0$, the \texttt{No Growth} phase will be skipped and $\gamma$ will be the proportion of pruned weights at the current pruning cycle. The parameters $\epsilon$ and $\delta$ determine the range of $\texttt{max\_lr}$.
As the network is iteratively pruned, $\gamma$ increases and the $\texttt{max\_lr}$ increases from $\epsilon$ to $\epsilon + \delta$ accordingly. The details are summarized in Algorithm \ref{algorithm1} in Fig. \ref{SC}.

\noindent {\bf Parameter Selection for S-Cyc: } Algorithm \ref{algorithm1} requires several inputs to implement the S-Cyc algorithm. The value of $\epsilon$ can be tuned using the validation accuracy of the unpruned network while the value of $\delta$ can be tuned using the validation accuracy of the pruned network with targeted sparsity. The pruning rate \texttt{p} and pruning cycles \texttt{L} are chosen to meet the target sparsity. The number of training epochs \texttt{t} should be large enough to guarantee the network convergence. Lastly, based on our experience, the value of $\texttt{q}$ and $\beta$ could be tuned in the range of [0, 3] and [3, 6], respectively.

\begin{table*}[!ht]
\centering
\setlength\tabcolsep{8.8pt}
\begin{tabular}{ll}
\toprule
\toprule
{\bf Schedule} & {\bf Description } (Iters: Iterations) \\ \midrule
constant LR (\texttt{a}) &  keep LR as \texttt{a} over all Iters and pruning cycles. \\
LR decay (\texttt{a}, \texttt{b}) & linearly decay the value of LR from \texttt{a} over \texttt{b} Iters.    \\
cyclical LR (\texttt{a}, \texttt{b}, \texttt{c}) & linearly vary between \texttt{a} and \texttt{b} with a step size of \texttt{c} Iters. \\
LR warmup (\texttt{a}, \texttt{b}, \texttt{c}, \texttt{d}, \texttt{e}) & linearly increase LR to \texttt{a} over \texttt{b} Iters, 10x drop at \texttt{c}, \texttt{d}, \texttt{e} Iters. \\ \midrule
S-Cyc ($\epsilon$, $\delta$, \texttt{q}, $\beta$, \texttt{b}, \texttt{c}, \texttt{d}, \texttt{e}) & (1) use $\epsilon$, $\delta$, \texttt{q}, $\beta$ to compute $\texttt{max\_lr}$. (2) do LR warmup ($\texttt{max\_lr}$, \texttt{b}, \texttt{c}, \texttt{d}, \texttt{e}) \\
\bottomrule
\bottomrule
\end{tabular}
\vspace{-3mm}
\caption{Descriptions of LR benchmarks and the S-Cyc LR schedule used in Experiments.}
\label{schedules}
\end{table*}

{\renewcommand{\arraystretch}{0.9}
\begin{table*}[!ht]
\centering
\setlength\tabcolsep{6.5pt}
\begin{tabular}{l rrrrrr}
\toprule
\toprule
{\small {\bf Config}: ResNet-20, CIFAR-10, Global Gradient} & {\small {\bf Params}: 227K} & {\small {\bf Train Steps}: 63K Iters} & {\small {\bf Batch}: 128} & {\small {\bf Pruning Rate}: 0.2}\\
\bottomrule
\end{tabular}
\setlength\tabcolsep{7.3pt}
\begin{tikzpicture}
\node (tab1) {%
\hspace*{0mm}\begin{tabular}{lccccccccc}
\centering
{\small Percent of Weights Remaining, $\lambda$} & 100 & 32.9 & 21.1  &  10.9 & 5.72 & 2.03 \\ \toprule
{\small (1) constant LR } ({\footnotesize 1\texttt{e}-2}) &  89.4$\pm${\scriptsize 0.4} & 86.5$\pm${\scriptsize 0.9} & 85.0$\pm${\scriptsize 0.7} & 82.5$\pm${\scriptsize 0.3}  & 78.9$\pm${\scriptsize 0.8} & 66.2$\pm${\scriptsize 1.3}  \\
{\small (2) LR decay} ({\footnotesize 2\texttt{e}-2, 63K}) & 90.0$\pm${\scriptsize 0.4} & 87.0$\pm${\scriptsize 0.8} & 85.7$\pm${\scriptsize 0.6} & 82.9$\pm${\scriptsize 0.6} & 79.6$\pm${\scriptsize 0.7} & 66.9$\pm${\scriptsize 1.9}  \\
{\small (3) cyclical LR} ({\footnotesize 0, 2.5\texttt{e}-2, 8K}) &  89.6$\pm${\scriptsize 0.3} & 87.2$\pm${\scriptsize 0.6} & 85.6$\pm${\scriptsize 0.5} & 83.0$\pm${\scriptsize 0.4} & 80.1$\pm${\scriptsize 0.8} & 67.8$\pm${\scriptsize 1.4} \\
{\small (4) LR-warmup-1} ({\footnotesize 3\texttt{e}-2, 20K, 20K, 25K, Nil}) & {\bf 90.1$\pm${\scriptsize 0.5}} & 87.4$\pm${\scriptsize 0.8} & 85.5$\pm${\scriptsize 0.6} & 83.4$\pm${\scriptsize 0.6} & 80.0$\pm${\scriptsize 1.4} & 68.9$\pm${\scriptsize 1.3} \\
{\small (5) LR-warmup-2} ({\footnotesize 7\texttt{e}-2, 20K, 20K, 25K, Nil}) & 89.7$\pm${\scriptsize 0.4} & 87.6$\pm${\scriptsize 0.5} & 85.9$\pm${\scriptsize 0.4} & 83.8$\pm${\scriptsize 0.8} & 80.6$\pm${\scriptsize 1.2} & 71.3$\pm${\scriptsize 1.2} \\
{\small (6) S-Cyc} ({\footnotesize 3\texttt{e}-2, 4\texttt{e}-2, 1, 5, 20K, 20K, 25K, Nil}) & 90.0$\pm${\scriptsize 0.3} & {\bf 88.3$\pm${\scriptsize 0.4}} & {\bf 87.1$\pm${\scriptsize 0.5}} & {\bf 84.5$\pm${\scriptsize 0.8}} & {\bf 81.7$\pm${\scriptsize 0.9}} & {\bf 72.8$\pm${\scriptsize 1.6}} \\
\bottomrule
\end{tabular}};
\node [right=of tab1,xshift = -13cm, yshift=0.6cm, ,opacity=.2,text opacity=.3] {\Huge (a)};
\end{tikzpicture}

\vspace{-3mm}
\end{table*}}

{\renewcommand{\arraystretch}{0.9}
\begin{table*}[!ht]
\centering
\setlength\tabcolsep{7.3pt}
\begin{tabular}{l rrrrr}
\toprule
{\small {\bf Config}: VGG-19, CIFAR-10, Layer Gradient} & {\small {\bf Params}: 139M} & {\small {\bf Train Steps}: 63K Iters} & {\small {\bf Batch}: 128} & {\small {\bf Pruning Rate}: 0.2} \\
\bottomrule
\end{tabular}
\setlength\tabcolsep{7.3pt}
\begin{tikzpicture}
\node (tab1) {%
\hspace*{1mm}\begin{tabular}{lcccccccc}
{\small Percent of Weights Remaining, $\lambda$} & 100 & 32.9 & 21.1  &  10.9 & 5.72 & 2.03 \\ \toprule
{\small (1) constant LR} ({\footnotesize 8\texttt{e}-3}) & 92.0$\pm${\scriptsize 0.3} & 87.9$\pm${\scriptsize 0.7}& 85.6$\pm${\scriptsize 0.9} & 80.2$\pm${\scriptsize 1.4} & 70.1$\pm${\scriptsize 1.3} & 40.1$\pm${\scriptsize 1.4}  \\
{\small (2) LR decay} ({\footnotesize 1\texttt{e}-2, 63K}) & 92.1$\pm${\scriptsize 0.5} & 88.4$\pm${\scriptsize 0.5} & 86.1$\pm${\scriptsize 0.6} & 80.3$\pm${\scriptsize 0.8} & 69.8$\pm${\scriptsize 1.9} & 40.9$\pm${\scriptsize 2.3}  \\
{\small (3) cyclical LR} ({\footnotesize 0, 3\texttt{e}-2, 15K})  & 92.3$\pm${\scriptsize 0.6} & 89.2$\pm${\scriptsize 0.4} & 87.2$\pm${\scriptsize 0.8} & 81.1$\pm${\scriptsize 1.3} & 70.4$\pm${\scriptsize 1.6} & 43.2$\pm${\scriptsize 1.9} \\
{\small (4) LR-warmup-1} ({\footnotesize 1\texttt{e}-1, 10K, 32K, 48K, Nil}) & 92.2$\pm${\scriptsize 0.3} & 89.4$\pm${\scriptsize 0.5} & 87.5$\pm${\scriptsize 0.7} & 81.7$\pm${\scriptsize 1.1} & 71.5$\pm${\scriptsize 1.3} & 46.6$\pm${\scriptsize 2.7} \\
{\small (5) LR-warmup-2} ({\footnotesize 4\texttt{e}-2, 10K, 32K, 48K, Nil}) & 92.9$\pm${\scriptsize 0.3} & 89.5$\pm${\scriptsize 0.4} & 87.0$\pm${\scriptsize 0.5} & 81.4$\pm${\scriptsize 1.2} & 70.9$\pm${\scriptsize 1.4} & 45.0$\pm${\scriptsize 2.3} \\
{\small (6) S-Cyc} ({\footnotesize 4\texttt{e}-2, 6\texttt{e}-2, 1, 4, 10K, 32K, 48K, Nil})   & {\bf 93.0$\pm${\scriptsize 0.4}} & {\bf 90.2$\pm${\scriptsize 0.7}} & {\bf 88.2$\pm${\scriptsize 1.4}} & {\bf 83.1$\pm${\scriptsize 1.1}} &{\bf 73.2$\pm${\scriptsize 0.9}} & {\bf 48.2$\pm${\scriptsize 2.1}} \\
\bottomrule
\end{tabular}};
\node [right=of tab1,xshift = -13cm, yshift=0.6cm, ,opacity=.2,text opacity=.3] {\Huge (b)};
\end{tikzpicture}
\vspace{-3mm}
\end{table*}}

{\renewcommand{\arraystretch}{0.9}
\begin{table*}[!ht]
\centering
\setlength\tabcolsep{8.9pt}
\begin{tabular}{l rrrrr}
\toprule
{\small {\bf Config}: ResNet-50, ImageNet-200, IMP} & {\small {\bf Params}: 23.8M} & {\small {\bf Train Steps}: 70K Iters} & {\small {\bf Batch}: 128} & {\small {\bf Pruning Rate}: 0.2} \\
\bottomrule
\end{tabular}
\setlength\tabcolsep{7.3pt}
\begin{tikzpicture}
\node (tab1) {%
\hspace*{1mm}\begin{tabular}{lcccccccc}
{\small Percent of Weights Remaining, $\lambda$} & 100 & 32.9 & 21.1  &  10.9 & 5.72 & 2.03 \\ \toprule
{\small (1) constant LR} ({\footnotesize 1\texttt{e}-2}) & 56.7$\pm${\scriptsize 0.3} & 54.4$\pm${\scriptsize 0.4} & 52.7$\pm${\scriptsize 1.2} & 50.1$\pm${\scriptsize 1.3} & 46.3$\pm${\scriptsize 0.9} & 43.9$\pm${\scriptsize 1.2} \\
{\small (2) LR decay} ({\footnotesize 3\texttt{e}-2, 70K)} & 56.7$\pm${\scriptsize 0.2} & 55.0$\pm${\scriptsize 0.5} & 53.1$\pm${\scriptsize 1.0} & 50.9$\pm${\scriptsize 1.2}  & 47.0$\pm${\scriptsize 0.7} & 41.8$\pm${\scriptsize 1.4}  \\
{\small (3) cyclical LR} ({\footnotesize 0, 5\texttt{e}-2, 20K}) & 56.6$\pm${\scriptsize 0.2} & 55.4$\pm${\scriptsize 0.4} & 54.3$\pm${\scriptsize 0.8} & 52.2$\pm${\scriptsize 1.1} & 46.0$\pm${\scriptsize 1.4} & 43.3$\pm${\scriptsize 1.2} \\
{\small (4) LR-warmup-1} ({\footnotesize 1\texttt{e}-1, 4K, 23K, 46K, 62K}) & 57.0$\pm${\scriptsize 0.3} & 56.2$\pm${\scriptsize 0.5} & 56.3$\pm${\scriptsize 0.6} & 53.7$\pm${\scriptsize 0.9} & 49.7$\pm${\scriptsize 1.1} & 47.4$\pm${\scriptsize 1.1} \\
{\small (5) LR-warmup-2} ({\footnotesize 5\texttt{e}-2, 4K, 23K, 46K, 62K}) & {\bf 57.8$\pm${\scriptsize 0.3}} & 56.6$\pm${\scriptsize 0.4} & 55.9$\pm${\scriptsize 0.4} & 53.3$\pm${\scriptsize 1.0} & 48.6$\pm${\scriptsize 1.0} & 45.0$\pm${\scriptsize 0.9} \\
{\small (6) S-Cyc} ({\footnotesize 5\texttt{e}-2, 5\texttt{e}-2, 2, 5, 4K, 23K, 46K, 62K}) &  57.6$\pm${\scriptsize 0.4} & {\bf 57.1$\pm${\scriptsize 0.5}} & {\bf 56.4$\pm${\scriptsize 0.6}} & {\bf 54.4$\pm${\scriptsize 1.3}} & {\bf 51.0$\pm${\scriptsize 1.2}} & {\bf 48.9$\pm${\scriptsize 1.5}} \\
\bottomrule
\bottomrule
\end{tabular}};
\node [right=of tab1,xshift = -13cm, yshift=0.65cm, ,opacity=.2,text opacity=.3] {\Huge (c)};
\end{tikzpicture}
\vspace{-4mm}
\caption{Performance comparison (averaged test accuracy $\pm$ std over 5 runs) of (a): pruning ResNet-20 on the CIFAR-10 dataset using global gradient; (b) pruning VGG-19 on the CIFAR-10 dataset using layer gradient; (c) pruning ResNet-50 on ImageNet-200 dataset using IMP. LR-warmup-1 is the standard implementation used in \cite{frankle2018lottery, frankle2020linear, renda2020comparing}. Results for more values of $\lambda$ are in Tables \ref{per1_extra} - \ref{per3_extra} in the \textbf{Appendix}.}
\label{per3}
\end{table*}}

\section{Performance Evaluation}
\label{PE}
We first summarize the experiment setup in Section \ref{ES} and compare the performance of S-Cyc to four benchmarks in Section \ref{PC}. In Section \ref{5.3}, we present the value of \texttt{max\_{lr}} estimated by S-Cyc at each pruning cycle and compare it to an oracle which carefully tunes $\texttt{max\_lr}$ via grid search.

\subsection{Experimental Setup}
\label{ES}
To demonstrate that S-Cyc can work well with different pruning methods, we shortlist three popular pruning methods (global weight, global gradient, layer gradient) suggested by \cite{Blalock2020} and one state-of-the-art pruning method (Iterative Magnitude Pruning (IMP)) \cite{frankle2018lottery}. The details are summarized as follows:
\begin{enumerate}[noitemsep,leftmargin=5mm, topsep=0pt]
\item Prune ResNet-20 on the CIFAR-10 dataset using global gradient (i.e., prune weights with the lowest absolute value of (weight $\times$ gradient) anywhere in the network).
\item Prune VGG-19 on the CIFAR-10 dataset using layer-wise gradient (i.e., prune weights with the lowest absolute value of (weight $\times$ gradient) in each layer).
\item Prune ResNet-50 on the ImageNet-200 dataset \cite{tinyimagenet} using IMP (i.e., prune weights with the lowest magnitude anywhere in the networks and then rewind the unpruned weights back to the initial values). 
\item The global magnitude is used to evaluate S-Cyc using different optimizers (see Section \ref{discussion}).
\end{enumerate}
In each experiment, we compare S-Cyc to constant LR and the three shortlisted LR schedules (discussed in Section \ref{sec4.1}): (i) LR decay, (ii) cyclical LR and (iii) LR warmup. The details of each LR schedule are  summarized in Table \ref{schedules}.

\noindent \textbf{Methodology:} In each run, the dataset is randomly split into three parts: training dataset (60\%), validation dataset (20\%) and testing dataset (20\%). We train the network using the training dataset with SGD and iteratively prune the trained network with a pruning rate of 0.2 (i.e., 20\% of remaining weights are pruned) in 1 pruning cycle. We repeat 25 pruning cycles in 1 run and use early-stop test accuracy (i.e., the corresponding test accuracy when early stopping criteria for validation error is met) to evaluate the performance. The results are averaged over 5 runs and the corresponding standard deviation are summarized in Table \ref{per3}. 

\noindent \textbf{Parameters for LR Warmup Schedule:} To ensure fair comparison against prior state-of-the-art, we utilize implementations reported in the literature, specifically, hyperparameters for ResNet-20 are from \cite{frankle2018lottery, frankle2020linear}, hyperparameters for VGG-19 are from \cite{frankle2018lottery, frankle2020linear, liu2018rethinking}, and hyperparameters for ResNet-50 are adapted from \cite{frankle2020linear, renda2020comparing}. We term this approach LR-warmup-1. Furthermore, to improve on the competitiveness of the state-of-the-art, we further tune the LR of LR-warmup-1 via a grid search from 1\texttt{e}-4 to 1\texttt{e}-1, which we term LR-warmup-2.

\noindent \textbf{Parameters for other LR schedules:} For the other schedules without a single "best" LR in the literature, we tune the value of LR for each of them via a grid search with range from 1\texttt{e}-4 to 1\texttt{e}-1 using the validation accuracy of the unpruned network. Other related parameters (e.g., step size of cyclical LR) are also tuned in the same manner. Lastly, we note that all benchmark LR schedules, including LR warmup, are rewound to the initial state at the beginning of every pruning cycle (same as the LR rewinding). 

\noindent \textbf{Source Code \& Devices:} We use NVIDIA RTX 2080 Ti devices for our experiments and the source code (including random seeds) will be released for reproducibility at the camera-ready stage (see Supplementary Material).

\subsection{Performance Comparison}
\label{PC}
In Table \ref{per3} (a), (b) \& (c), we observe that LR warmup schedule generally outperforms other benchmarks (i.e., compare rows 4 \& 5 to rows 1, 2 \& 3). This agrees with the results stated in prior works \cite{renda2020comparing}. 

\noindent \textbf{LR-warmup-1 vs LR-warmup-2:} In general, the results indicate that higher LRs lead to higher test accuracy in pruned networks, with slightly lower test accuracy in the unpruned network. As an example, in Table \ref{per3} (a), LR-warmup-1 has an maximal LR of 3$\texttt{e}$-2, which is lower than that of LR-warmup-2's 7$\texttt{e}$-2. As such, it can be seen that LR-warmup-1 achieves a higher test accuracy for the unpruned network ($\lambda = 100$), while achieving a lower test accuracy for the pruned network ($\lambda \leq 32.9$), as compared to LR-warmup-2. A similar finding can be obtained in Table \ref{per3} (b) \& (c), when LR-warmup-1 has a higher maximal LR than LR-warmup-2. These empirical results verify our new insight in Section \ref{sec3}, namely that a larger value of LR should be used as the neural network is iteratively pruned.

\begin{table*}[!t]
\centering
\setlength\tabcolsep{7.7pt}
\begin{tabular}{lcccccccccc}
\bottomrule
\bottomrule
{\small Percent of Weights Remaining, $\lambda$} & 100 & 64 & 51.3 & 41.1 & 32.9 & 21.1 & 5.72\\ \midrule
{\small Well-Tuned \texttt{max\_lr} ($\times$\texttt{e}-2)} & 4 & 4.2 & 4.6 & 6.2 & 9.0 & 9.8  & 10.2 \\
{\small Feasible LR Region ($\times$\texttt{e}-2)} & {\small [3.6, 4.2]} & {\small [3.6, 4.8]} & {\small [4.2, 5.4]} & {\small [5.6, 7.2]} & {\small [8.0, 9.6]} & {\small [9.2, 10.4]} & {\small [9.8, 10.6]} \\
{\small S-Cyc ($\times$\texttt{e}-2)} & 4 & 4.02 & 4.5 & 6.7 & 8.8 & 9.9 & 9.99 \\
\bottomrule
\bottomrule
\end{tabular}
\vspace{-3mm}
\caption{Comparison between the value of $\texttt{max\_lr}$ estimated by S-Cyc, well-tuned $\texttt{max\_lr}$ via grid search and feasible LR region when iteratively pruning VGG-19 on CIFAR-10 dataset \cite{cifar10} using the layer gradient pruning method \cite{Blalock2020}. Note that all values in the table are in hundredths.}
\vspace{-0mm}
\label{lr_table_vgg}
\end{table*}

\noindent \textbf{Reproducing state-of-the-art results:} By using the standard LR schedules, we have successfully reproduced the state-of-the-art results reported in the literature. For example, the performance of LR-warmup-1 in Table \ref{per3} (a) \& (b) are comparable to those reported in the literature (see the green line in Fig. 11 and the red line in Fig. 9 in \cite{Blalock2020}). Moreover, the performance of unpruned network is also an important performance indicator. The performance of unpruned network in Table \ref{per3} are compare to those reported in \cite{frankle2018lottery, tanaka2020pruning} (compare Table \ref{per3} (a), (b)  \& (c)  to Figs. 7,  8 in \cite{frankle2018lottery} and Fig.6 in \cite{tanaka2020pruning}, respectively).

\noindent \textbf{S-Cyc outperforms state-of-the-art results:} The performance of S-Cyc is comparable to LR-warmup-1 in Table \ref{per3} (a) and LR-warmup-2 in Tables \ref{per3} (b) \& (c) for the unpruned network. It is because S-Cyc uses the same LR at the first few pruning cycles. The key innovation of S-Cyc is that the LR increases as the network is pruned, by gradually increasing $\texttt{max\_lr}$ as $\lambda$ decreases. This results in a much higher accuracy than all LR benchmarks studied. For example, in Table \ref{per3} (a), the accuracy of S-Cyc is 2.1\% higher than the best performing schedule (LR-warmup-2) when $\lambda$ reduces to 2.03. S-Cyc also performs the best when using larger models in Table \ref{per3} (b) (i.e., 3.4\% higher at $\lambda$ = 2.03) and when using larger datasets and state-of-the-art pruning method (IMP) in Table \ref{per3} (c) (i.e., 3.2\% higher at $\lambda$ = 2.03).


\noindent \textbf{Complexity Analysis:} S-Cyc only introduces three additional tunable parameters to the LR warmup schedule (see Table \ref{schedules}) and those additional parameters such as $\beta$ and $\texttt{q}$ can be tuned within a very narrow range as suggested in Section \ref{sec4.3}, which does not cause substantial increase in complexity (i.e., approximately up to 20 additional trials).

\subsection{Comparing S-Cyc's to an Oracle}
\label{5.3}
Having established that S-Cyc outperforms other LR schedules, we evaluate if our S-Cyc's estimated \texttt{max\_{lr}} is competitive as compared to a greedy oracle which has a well-tuned \texttt{max\_{lr}} at each pruning cycle. The oracle's \texttt{max\_{lr}} at the current pruning cycle is chosen by grid search ranging from 1$\texttt{e}$-4 to 1$\texttt{e}$-1 and the best performing value (i.e. determined by validation accuracy) is used to train the network. The results of \texttt{max\_{lr}} determined this way when iteratively pruning a VGG-19 on CIFAR-10 using the layer gradient is detailed in Table \ref{lr_table_vgg} via two metrics:
\begin{enumerate}[noitemsep,leftmargin=5mm, topsep=0pt]
\item {\bf Well-tuned $\texttt{max\_lr}$}: The value of $\texttt{max\_lr}$ which provides the best validation accuracy. 
\item {\bf Feasible LR Region}: Range of $\texttt{max\_lr}$ which performs within 0.5\% of best validation accuracy.
\end{enumerate}
As can be seen in Table \ref{lr_table_vgg}, both the value of well-tuned $\texttt{max\_lr}$ and the upper/lower bound of the feasible LR region tend to increase as the network is iteratively pruned. This concurs with our new insight in Section \ref{sec3}. 

\noindent \textbf{Oracle vs S-Cyc:} The value of $\texttt{max\_lr}$ estimated by S-Cyc falls in the feasible LR region at each pruning cycle, meaning the validation accuracy of S-Cyc is comparable to the well-tuned $\texttt{max\_lr}$. We note that the process of finding the well-tuned $\texttt{max\_lr}$ requires a significantly larger amount of time in tuning due to the grid search. Using S-Cyc, we are able to obtain a comparable performance to the oracle, with exponentially less time, highlighting the competitiveness of S-Cyc. Similar results were also obtained using ResNet-20, global gradient pruning method on CIFAR-10 and ResNet-50, IMP on ImageNet-200 and we refer the interested reader to Tables \ref{lr_table_resnet_20} and \ref{lr_table_resnet_50} in the \textbf{Appendix}.

\section{Discussion}
\label{discussion}

{\bf Connection to Prior Work:} Our work explores the important role of LR in network pruning and provides a new insight -- {\em as the ReLU-based network is iteratively pruned, a larger LR should be used}. This new insight provides an explanation of several phenomena observed in prior works. Specifically, \citeauthor{frankle2018lottery} (2019) highlight that IMP is sensitive to the LR used and they can only find winning tickets after applying a LR warmup schedule. Using insights from our analysis, we attribute this to LR warmup increasing the LR to a large value (e.g., \cite{frankle2018lottery} increases LR to 1$\texttt{e}$-1 when training VGG-19) which is better for pruned networks. Similarly, \citeauthor{renda2020comparing} (2020) propose LR rewinding and demonstrate it outperforms standard fine-tuning. We attribute this to LR rewinding ensuring that a relatively larger LR is used as the network is pruned. Furthermore, we highlight that the proposed S-Cyc obtains better performance than LR rewinding. It also works well with IMP and achieves a higher accuracy.

\noindent {\bf Performance of S-Cyc using other Adaptive LR Optimizers: } In the main paper, we only evaluate the performance of S-Cyc using SGD. We note that the weight update mechanism is different for other adaptive learning rate optimizers, which may potentially affect the performance of S-Cyc. In Tables \ref{per_adam} - \ref{per_rms} in the \textbf{Appendix}, we conduct a similar performance comparison using Adam \cite{kingma2014adam} and RMSprop \cite{tieleman2012lecture}, and S-Cyc still outperforms all LR benchmarks studied. 



\noindent {\bf Future Research: } We evaluate the performance of S-Cyc using several state-of-the-art networks (e.g., ResNet-50) and popular datasets (e.g., ImageNet-200). We intend to explore the applicability of S-Cyc on larger datasets (e.g., ImageNet) in future research. Furthermore, we only demonstrate the performance of S-Cyc on ReLU-based networks. We note that similar LR schedules could be used for networks with other activation functions (e.g., PReLU). We will also explore this part in our future research. Lastly, the main motivation for S-Cyc is that the distribution of weight gradients tends to become narrower after pruning. Automatically determine the value of $\texttt{max\_lr}$ from the distribution of weight gradients could be the next topic to be explored.

\clearpage
\balance
\bibliography{sample-bibliography}

\clearpage
\onecolumn
\appendix
\section{Supplementary Results}
In Section \ref{A1}, we first show more results on the 
distribution of weight gradients. Next, we present the performance comparison using Adam \cite{kingma2014adam} and RMSProp \cite{tieleman2012lecture} optimizers in Section \ref{A2}. In Section \ref{A4}, we show the performance comparison between S-Cyc's $\texttt{max\_lr}$ to that of an oracle using ResNet-20, global gradient pruning method on CIFAR-10 and ResNet-50, IMP on ImageNet-200. Lastly, we show the results for more values of $\lambda$ in Section \ref{A3}

\subsection{More Experimental Results on the Distribution of Weight Gradients and Hidden Representations}
\label{A1}

\begin{figure}[!h]
\hspace{4mm}\begin{minipage}{0.45\textwidth}
\includegraphics[width=0.88\linewidth, left]{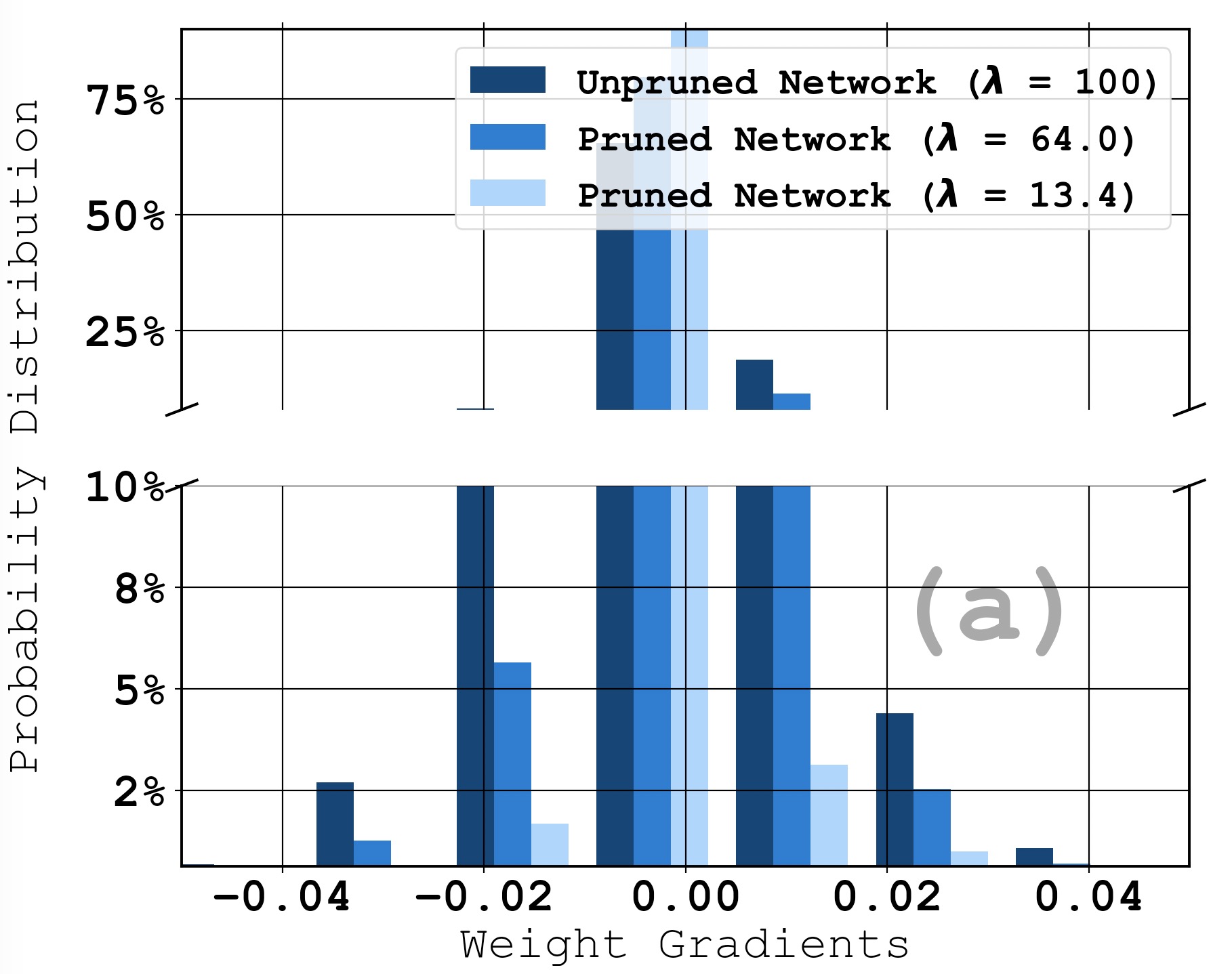}
\end{minipage}%
\hspace{-2mm}\begin{minipage}{0.45\textwidth}
\includegraphics[width=0.85\linewidth, right]{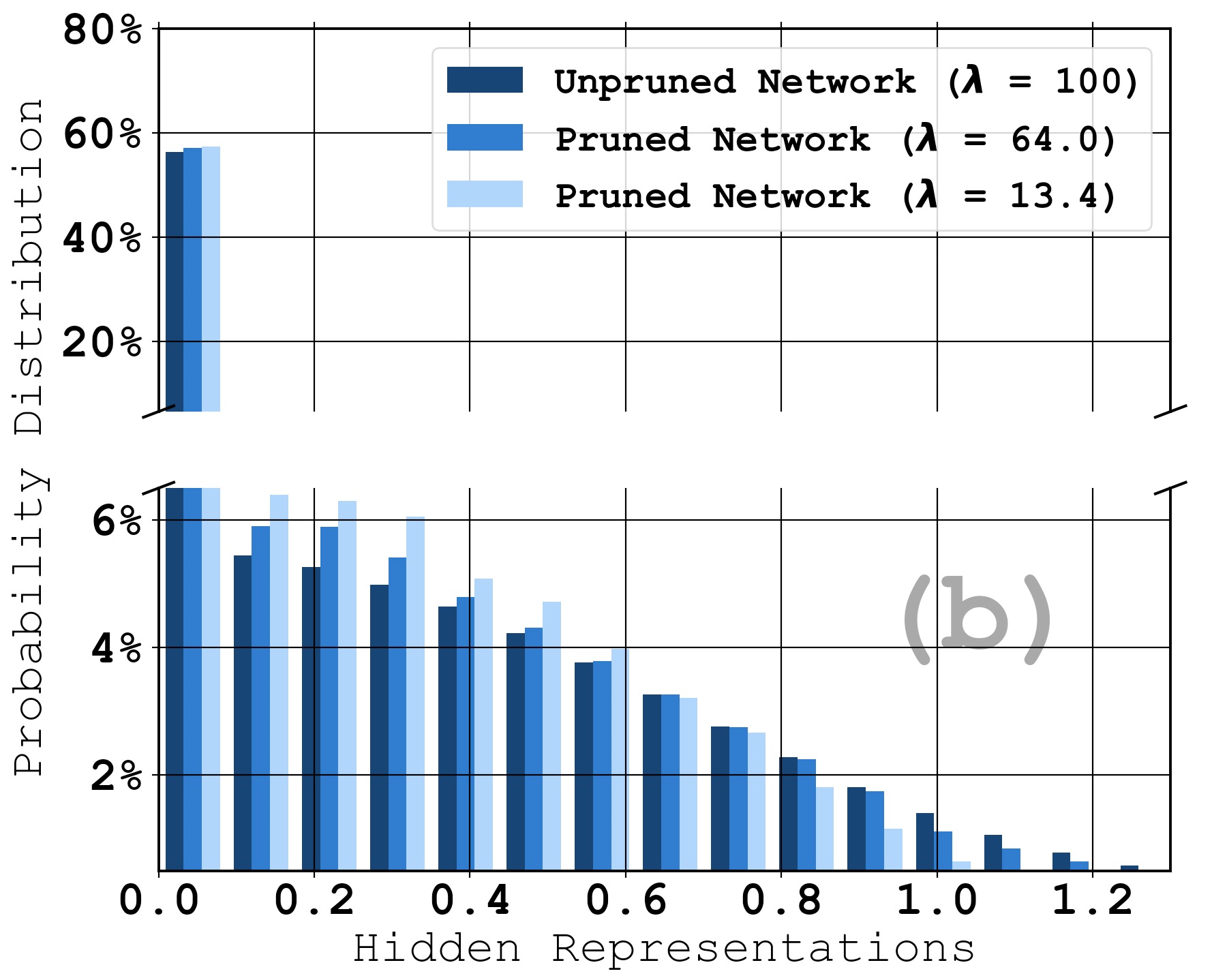}
\end{minipage}
\vspace{-3mm}
\caption{(a) Distribution of weight gradients when iteratively pruning a fully connected ReLU-based network using global magnitude \cite{han2015learning}, where $\lambda$ is the percent of weights remaining. (b) Corresponding distribution of hidden representations (i.e., post-activation outputs of all hidden layers). The difference with Fig. \ref{mlp_no_norm} is that the batch normalization technique \cite{ioffe2015batch} ($\mu$ = 0, $\sigma^2$ = 1) is used here.} 
\label{mlp_distribution_norm}
\end{figure}

\begin{figure}[!ht]
\hspace{4mm}\begin{minipage}{0.45\textwidth}
\includegraphics[width=0.9\linewidth, left]{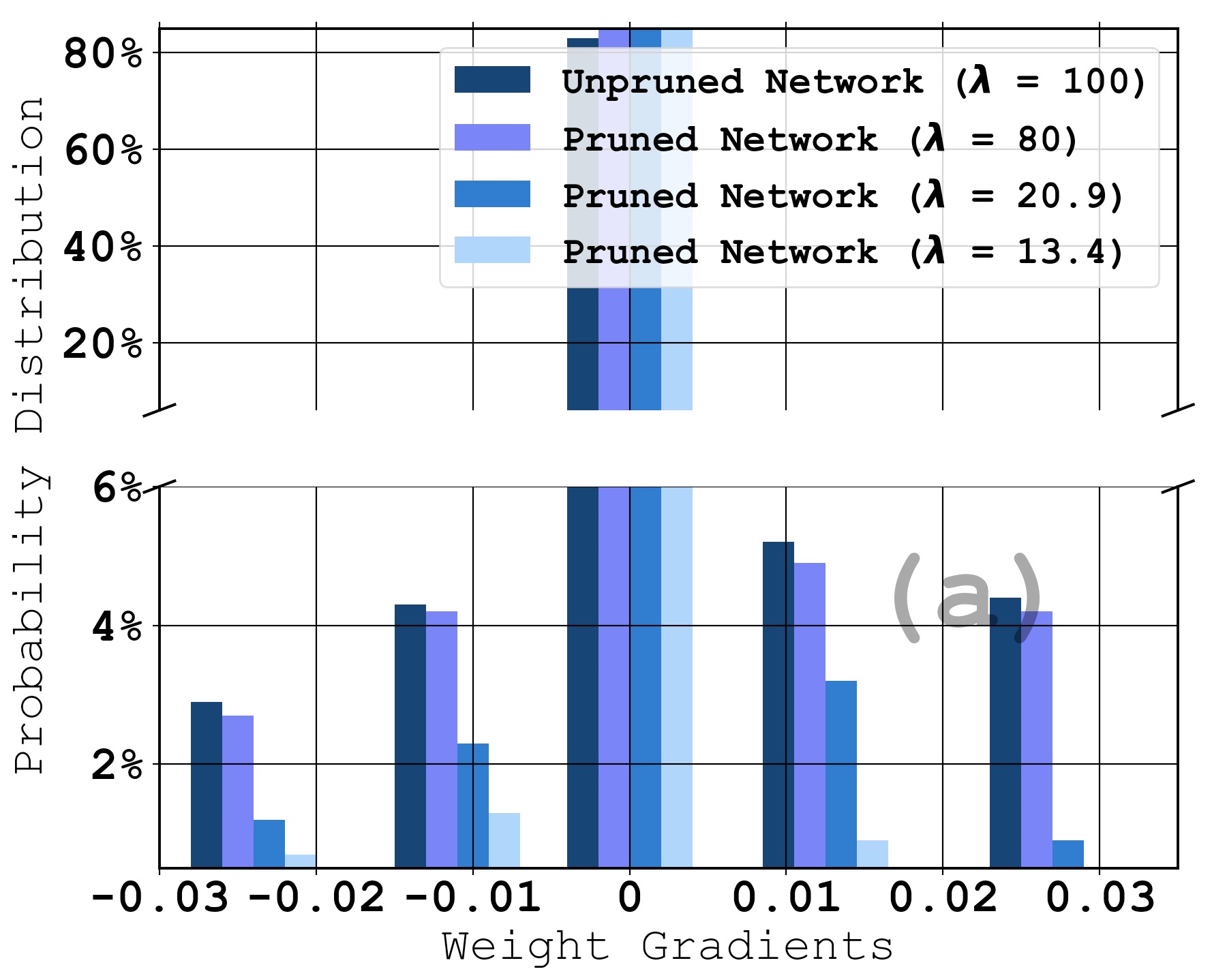}
\end{minipage}%
\hspace{-2mm}\begin{minipage}{0.45\textwidth}
\includegraphics[width=0.9\linewidth, right]{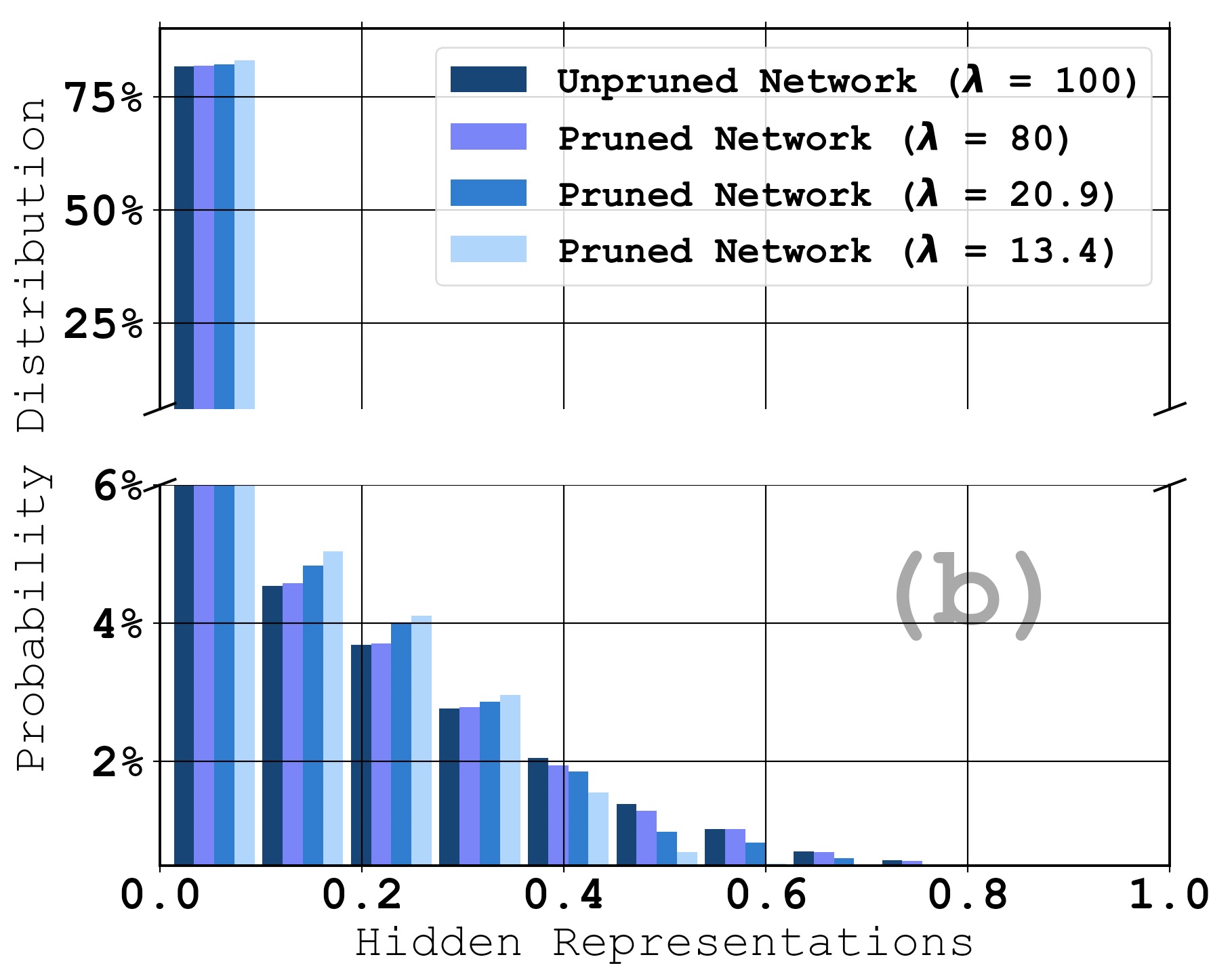}
\end{minipage}
\vspace{-3mm}
\caption{(a) Distribution of weight gradients when iteratively pruning the AlexNet network using the layer magnitude pruning method, where $\lambda$ is the percent of weights remaining. (b) Corresponding distribution of hidden representations (i.e., post-activation outputs of all hidden layers).} 
\label{alex}
\end{figure}

\begin{figure}[!ht]
\hspace{4mm}\begin{minipage}{0.45\textwidth}
\includegraphics[width=0.9\linewidth, left]{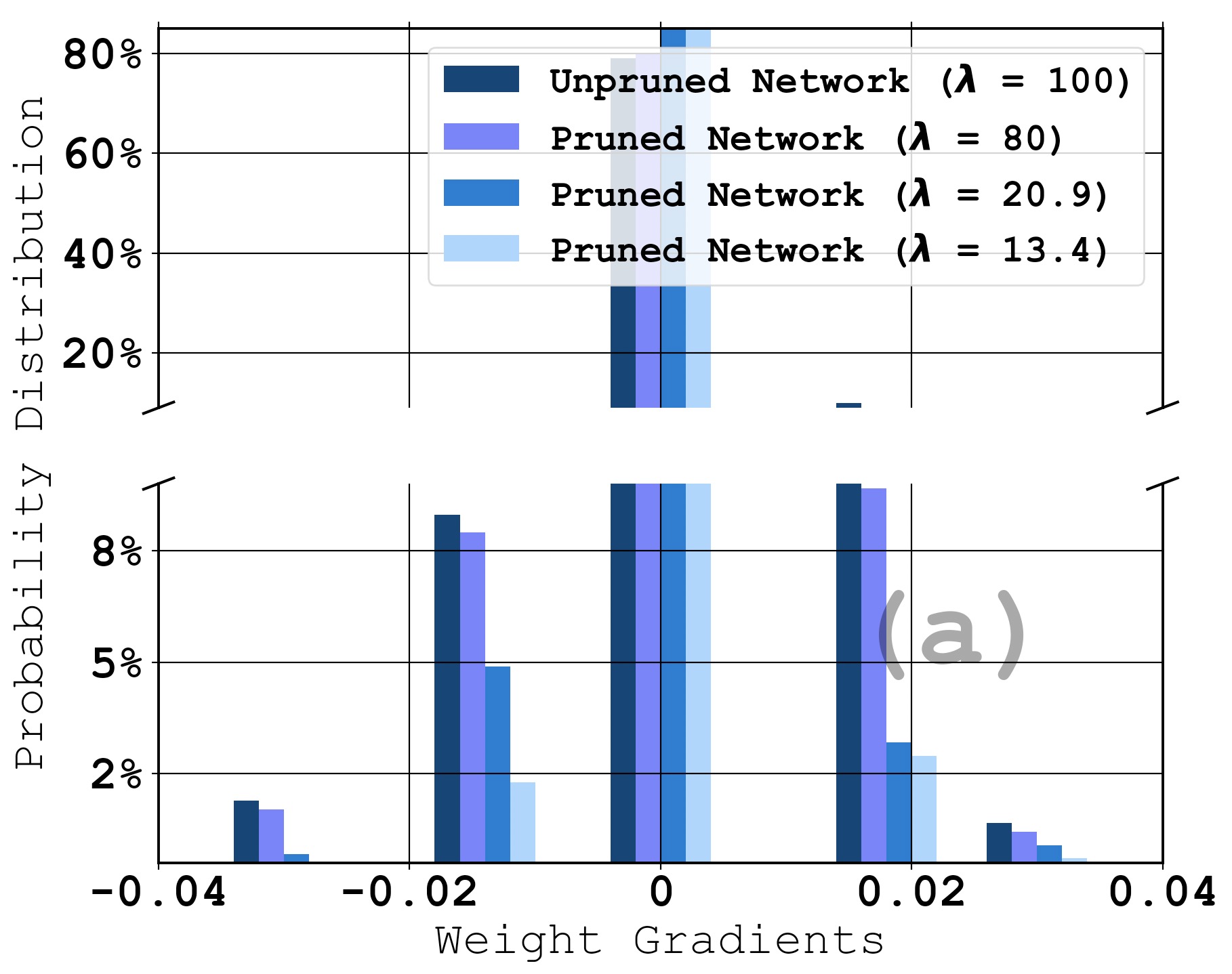}
\end{minipage}%
\hspace{-2mm}\begin{minipage}{0.45\textwidth}
\includegraphics[width=0.89\linewidth, right]{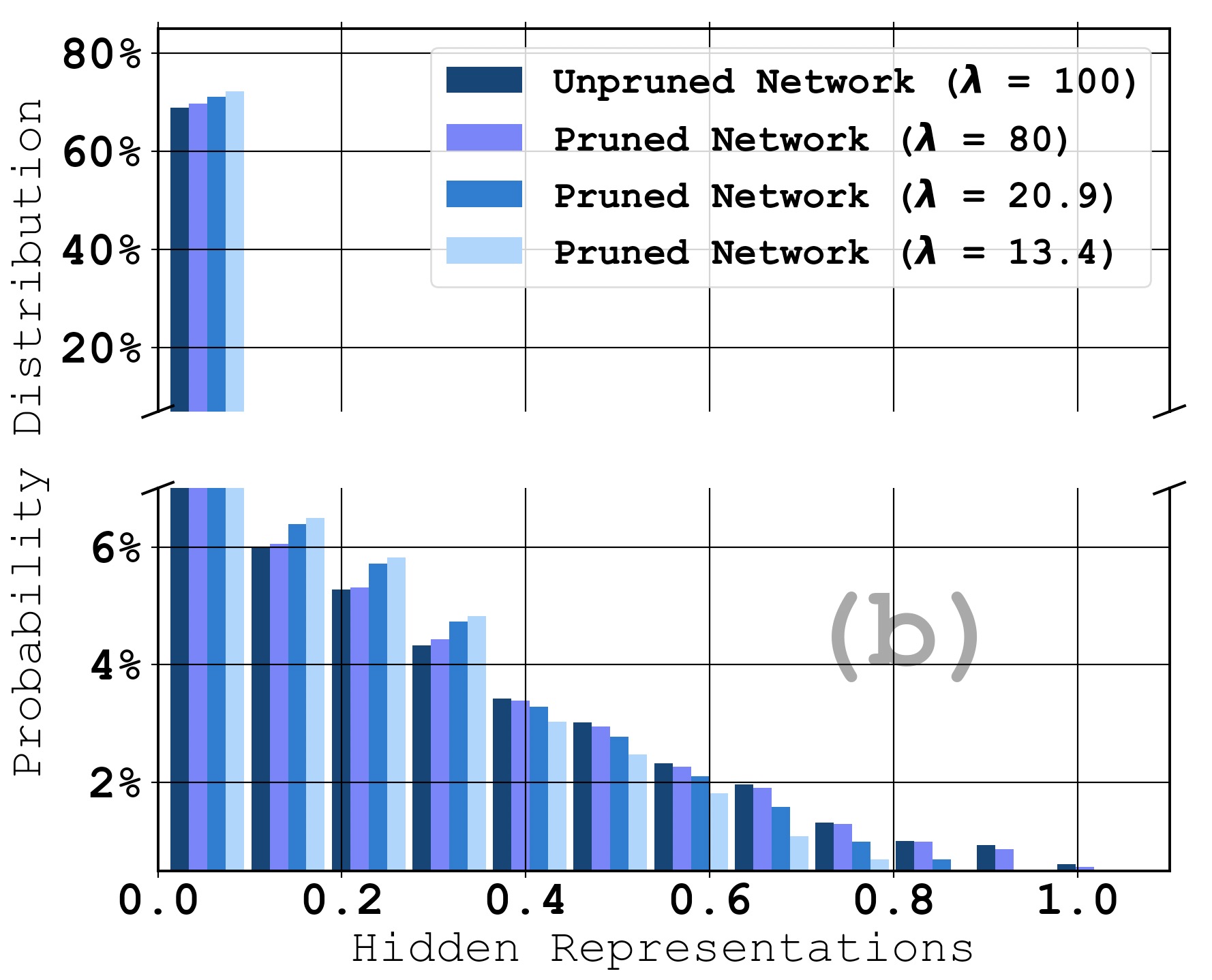}
\end{minipage}
\vspace{-3mm}
\caption{(a) Distribution of weight gradients when iteratively pruning the ResNet-20 network using the global gradient pruning method, where $\lambda$ is the percent of weights remaining. (b) Corresponding distribution of hidden representations (i.e., post-activation outputs of all hidden layers).} 
\label{resnet}
\end{figure}

\begin{figure}[!ht]
\hspace{2.5mm}\begin{minipage}{0.46\textwidth}
\includegraphics[width=0.9\linewidth, left]{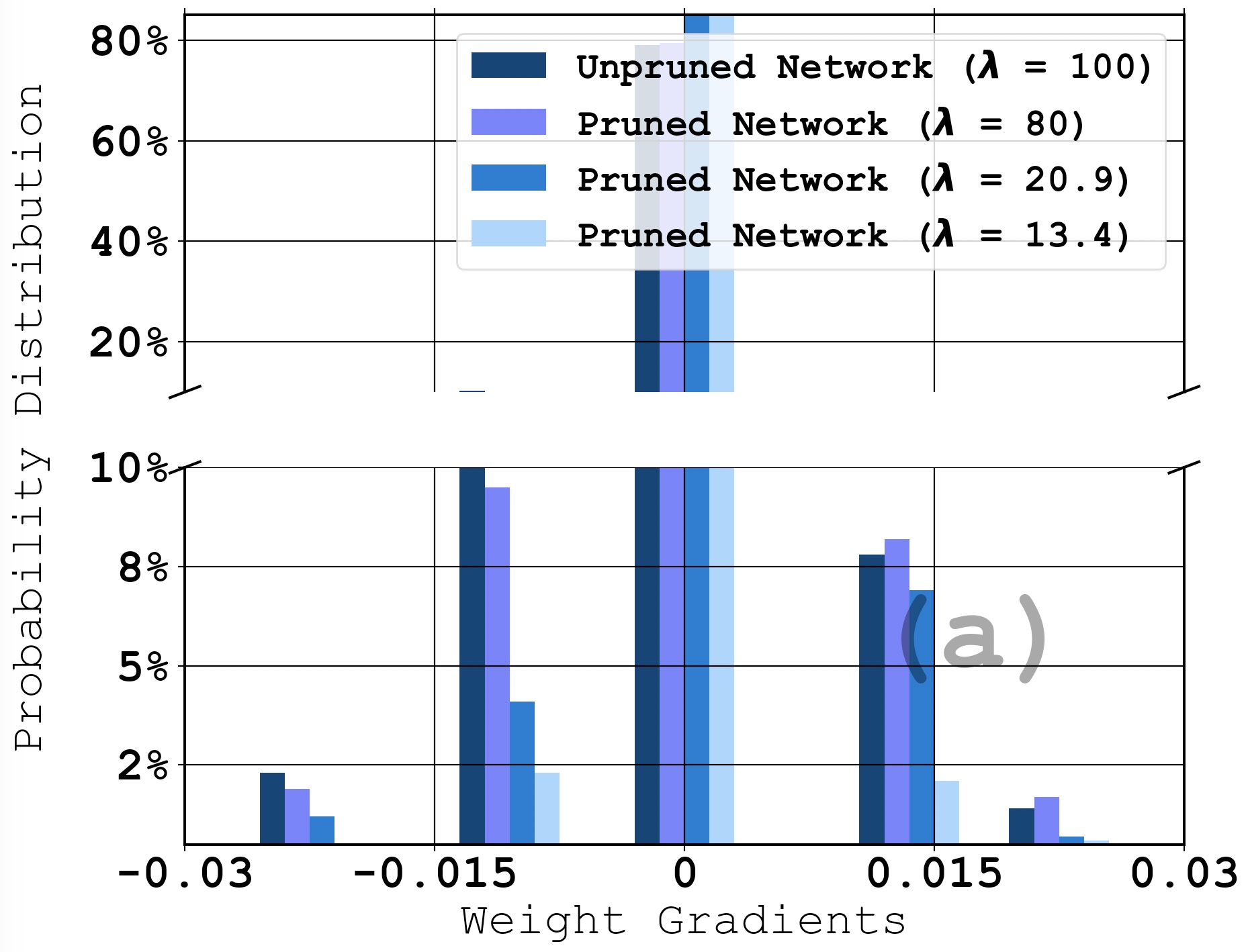}
\end{minipage}%
\hspace{-4mm}\begin{minipage}{0.46\textwidth}
\includegraphics[width=0.87\linewidth, right]{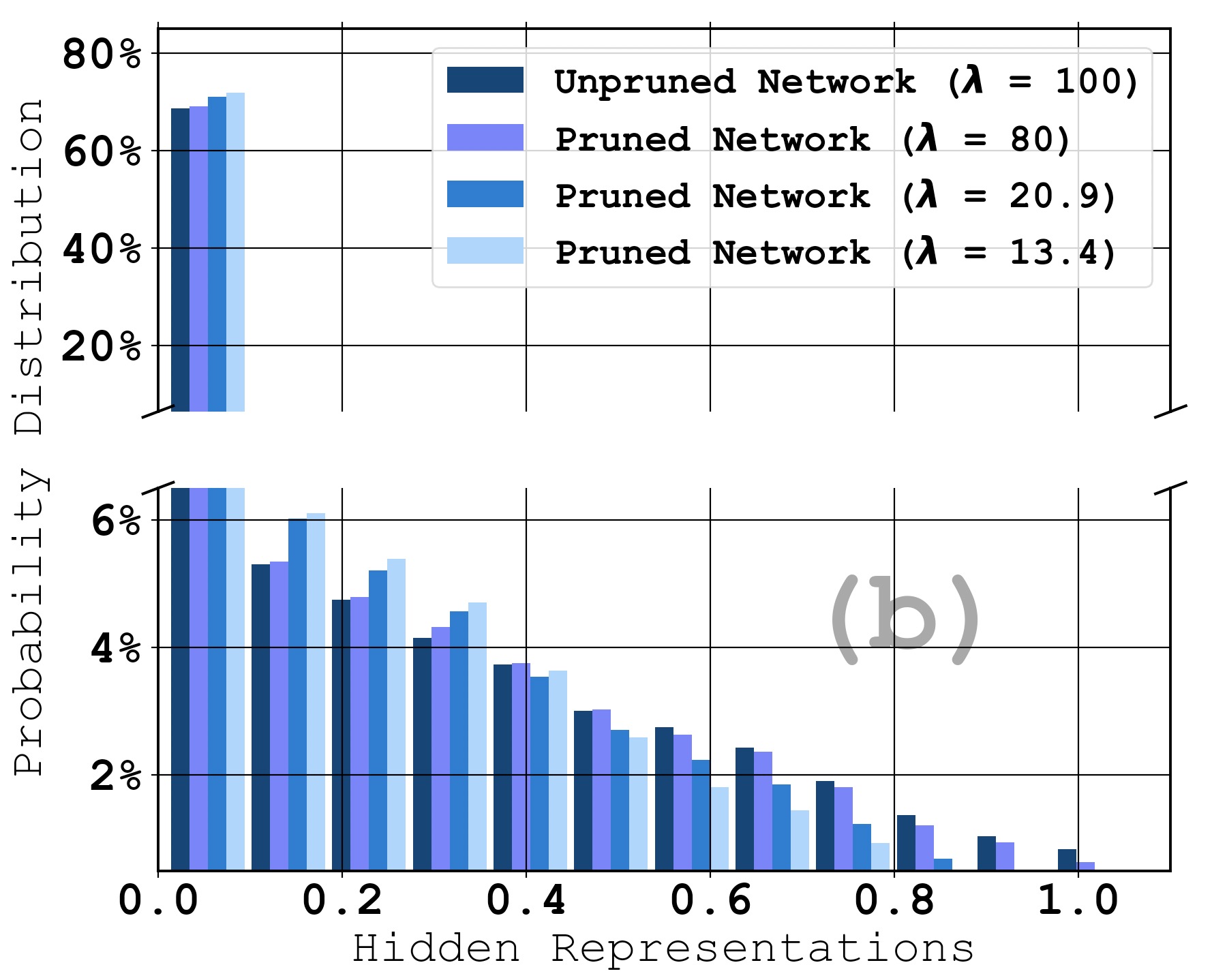}
\end{minipage}
\vspace{-3mm}
\caption{(a) Distribution of weight gradients when iteratively pruning the VGG-16 network using the structured filter pruning method \cite{li2016pruning}, where $\lambda$ is the percent of weights remaining. (b) Corresponding distribution of hidden representations (i.e., post-activation outputs of all hidden layers).} 
\label{vgg}
\end{figure}

In this subsection, we present more experimental results on the distribution of weight gradients and hidden representations using other datasets, networks and pruning methods in Figs. \ref{mlp_distribution_norm} - \ref{vgg}. The configuration for each network is given in Table \ref{arch_sec2}. 

We observe that the experimental results in Figs. \ref{mlp_distribution_norm} - \ref{vgg} largely mirror those in Fig. \ref{mlp_no_norm} (a). Specifically, in Fig.\ref{mlp_distribution_norm}, we show the distribution of weight gradients and hidden representations when iterative pruning a fully connected ReLU-based network using global magnitude with batch normalization applied to each hidden layer. We observe that the unpruned network ($\lambda$ = 100) has more than 6\% of weight gradients with values greater than 0.018 (the rightmost 2 bars) or less than -0.025 (the leftmost bar), while the pruned network ($\lambda$ = 13.4) has less than 1\% of weight gradients falling into those regions (see Fig.\ref{mlp_distribution_norm} (a)). Similarly, in Fig. \ref{alex} (a), the unpruned network ($\lambda$ = 100) has more than 7\% of weight gradients with values greater than 0.02 (the rightmost bar) or less than -0.02 (the left most bar), while the pruned network ($\lambda$ = 13.4) has less than 2\% of weight gradients falling into those regions. In Fig. \ref{resnet} (a), the unpruned network ($\lambda$ = 100) has more than 2\% of weight gradients with values greater than 0.025 (the rightmost bar) or less than -0.02 (the left most bar), while the pruned network ($\lambda$ = 13.4) has almost 0\% of weight gradients falling into those regions. In Fig. \ref{vgg} (a), the unpruned network ($\lambda$ = 100) has more than 3\% of weight gradients with values greater than 0.018 (the rightmost bar) or less than -0.015 (the left most bar), while the pruned network ($\lambda$ = 13.4) has almost 0\% of weight gradients falling into those regions. Furthermore, the corresponding distributions of hidden representations are also shown, which largely mirror those in Fig. \ref{mlp_no_norm} (b).

{\renewcommand{\arraystretch}{1.2}
\begin{table}[!ht]
\small
\setlength\tabcolsep{5.0pt} 
\centering
  \begin{tabular}{lc|cccccccc} \hline \hline
{Network}    & {Params} & {Train Steps} & {Batch} & {Optimizer} & {Learning Rate Schedule} & {BatchNorm} & {Metric} \\ \hline
AlexNet       & 57M         &  781K Iterations          &  64    &  SGD      & 0 to 1\texttt{e}-2 warmup over 150K              &  No                &  Layer Weight\\
                   &                 &                            &             &                & 10x drop at 300K, 400K                       &               &\\
ResNet-20  & 274K       &  63K Iterations            &128       &   SGD     & 0 to 3\texttt{e}-2 warmup over 20K, & Yes      & Global Gradient\\
                   &                &                            &             &                &     10x drop at 20K, 25K                   &               &                 \\
VGG-16      & 134M      &   63K Iterations           & 128       &  SGD       & 0 to 1\texttt{e}-1  warmup over 10K, & Yes      & L1 Norm\\
                   &                &                            &             &                &     10x drop at 32K, 48K                   &               & \\ \hline \hline             
  \end{tabular}
  \vspace{-3mm}
  \caption{Architectures and training details used in the appendix.}
  \vspace{-0mm}
  \label{arch_sec2}
\end{table}}

\newpage

\subsection{Performance Comparison using Adam and RMSProp}
\label{A2}

In this subsection, we show the performance comparison between the proposed S-Cyc and selected LR benchmarks using Adam \cite{kingma2014adam} and RMSProp \cite{tieleman2012lecture} optimizers. The experimental results summarized in Tables \ref{per_adam} -  \ref{per_rms} largely mirror those in Table \ref{per3}. Specifically, the proposed S-Cyc outperforms the best performing benchmark by a range of 0.8\% -2.7\% in pruned networks. 

{\renewcommand{\arraystretch}{1.1}
\begin{table}[!ht]
\centering
\setlength\tabcolsep{11pt}
\begin{tabular}{l rrrrr}
\toprule
\toprule
{\small {\bf Params}: 227K} & {\small {\bf Train Steps}: 63K Iters} & {\small {\bf Batch}: 128} & {\small {\bf Batch Norm}: Yes} & {\small {\bf Optimizer}: Adam} & {\small {\bf Rate}: 0.2}\\
\bottomrule
\end{tabular}

\setlength\tabcolsep{7.6pt}
\hspace*{1mm}\begin{tabular}{lcccccccccc}
{\small Percent of Weights Remaining, $\lambda$} & 100 & 32.9 & 21.1 & 10.9 & 5.72 & 2.03 \\ \midrule
{\small constant LR} ({\footnotesize 8\texttt{e}-4}) & 88.4$\pm${\scriptsize 0.4} & 84.8$\pm${\scriptsize 0.6} & 83.5$\pm${\scriptsize 0.6} & 80.5$\pm${\scriptsize 0.9} & 75.5$\pm${\scriptsize 1.2} & 67.1$\pm${\scriptsize 1.7} \\
{\small LR decay} ({\footnotesize 3\texttt{e}-3, 63K}) & 88.6$\pm${\scriptsize 0.3} & 87.1$\pm${\scriptsize 0.7} & 83.7$\pm${\scriptsize 0.9} & 82.1$\pm${\scriptsize 1.1} & 76.1$\pm${\scriptsize 0.8} & 66.0$\pm${\scriptsize 1.3} \\
{\small cyclical LR} ({\footnotesize 0, 3\texttt{e}-2, 8K})  & 88.9$\pm${\scriptsize 0.3} & 86.9$\pm${\scriptsize 0.5} & 84.1$\pm${\scriptsize 0.3} & 82.1$\pm${\scriptsize 0.5} & 77.0$\pm${\scriptsize 0.9} & 64.4$\pm${\scriptsize 1.1} \\
{\small LR-warmup-1} ({\footnotesize 3\texttt{e}-3, 20K, 20K, 25K, Nil}) & 89.1$\pm${\scriptsize 0.3} & 87.2$\pm${\scriptsize 0.4} & 84.5$\pm${\scriptsize 0.6} & 82.6$\pm${\scriptsize 0.7} & 75.2$\pm${\scriptsize 1.1} & 65.1$\pm${\scriptsize 1.9} \\
{\small LR-warmup-2} ({\footnotesize 7\texttt{e}-3, 20K, 20K, 25K, Nil}) & 87.8$\pm${\scriptsize 0.3} & 85.7$\pm${\scriptsize 0.5} & 84.3$\pm${\scriptsize 0.7} & 82.9$\pm${\scriptsize 0.8} & 78.1$\pm${\scriptsize 1.4} & 69.8$\pm${\scriptsize 2.1} \\
{\small S-Cyc} ({\footnotesize 3\texttt{e}-3, 4\texttt{e}-3, 1, 4, 20K, 20K, 25K, Nil})   &  {\bf 89.2$\pm${\scriptsize 0.2}} & {\bf 87.9$\pm${\scriptsize 0.3}} & {\bf 86.3$\pm${\scriptsize 0.5}} & {\bf 84.4$\pm${\scriptsize 0.6}} & {\bf 79.5$\pm${\scriptsize 1.7}} & {\bf 71.7$\pm${\scriptsize 2.3}} \\
\bottomrule
\bottomrule
\end{tabular}
\vspace{-3mm}
\caption{Performance comparison (averaged test accuracy $\pm$ std over 5 runs) of iteratively pruning ResNet-20 on CIFAR-10 dataset using the global weight magnitude pruning method \cite{han2015learning} and the Adam optimizer \cite{kingma2014adam}.}
\label{per_adam}
\end{table}}

{\renewcommand{\arraystretch}{1.1}
\begin{table}[!ht]
\centering
\setlength\tabcolsep{10.0pt}
\begin{tabular}{l rrrrr}
\toprule
\toprule
{\small {\bf Params}: 227K} & {\small {\bf Train Steps}: 63K Iters} & {\small {\bf Batch}: 128} & {\small {\bf Batch Norm}: Yes} & {\small {\bf Optimizer}: RMSProp} & {\small {\bf Rate}: 0.2}\\
\bottomrule
\end{tabular}

\setlength\tabcolsep{7.7pt}
\hspace*{0.6mm}\begin{tabular}{lcccccccccc}
{\small Percent of Weights Remaining, $\lambda$} & 100 & 32.9 & 21.1 & 10.9 & 5.72 & 2.03 \\ \midrule
{\small constant LR} ({\footnotesize 6\texttt{e}-4}) & 87.9$\pm${\scriptsize 0.3} & 83.4$\pm${\scriptsize 0.4} & 81.5$\pm${\scriptsize 0.9} & 77.0$\pm${\scriptsize 0.8} & 65.5$\pm${\scriptsize 1.9} & 55.1$\pm${\scriptsize 2.3} \\
{\small LR decay} ({\footnotesize 2\texttt{e}-3, 63K}) & 88.4$\pm${\scriptsize 0.2} & 84.8$\pm${\scriptsize 0.6} & 81.3$\pm${\scriptsize 1.1} & 77.8$\pm${\scriptsize 0.9} & 67.1$\pm${\scriptsize 1.4} & 58.3$\pm${\scriptsize 1.6} \\
{\small cyclical LR} ({\footnotesize 0, 3\texttt{e}-3, 10K})  & 88.1$\pm${\scriptsize 0.3} & 84.7$\pm${\scriptsize 0.5} & 81.9$\pm${\scriptsize 0.7} & 78.1$\pm${\scriptsize 1.0} & 67.5$\pm${\scriptsize 0.9} & 56.3$\pm${\scriptsize 1.7}   \\
{\small LR-warmup-1} ({\footnotesize 1\texttt{e}-3, 20K, 20K, 25K, Nil}) & {\bf 88.9$\pm${\scriptsize 0.2}} & 85.1$\pm${\scriptsize 0.5} & 81.7$\pm${\scriptsize 0.4} & 78.6$\pm${\scriptsize 0.6} & 67.3$\pm${\scriptsize 1.3} & 57.1$\pm${\scriptsize 1.4} \\
{\small LR-warmup-2} ({\footnotesize 3\texttt{e}-3, 20K, 20K, 25K, Nil}) & 88.5$\pm${\scriptsize 0.4} & 85.8$\pm${\scriptsize 0.6} & 82.6$\pm${\scriptsize 0.5} & 79.9$\pm${\scriptsize 0.7} & 69.3$\pm${\scriptsize 1.1} & 59.2$\pm${\scriptsize 2.3} \\
{\small S-Cyc} ({\footnotesize 1\texttt{e}-3, 2\texttt{e}-3, 2, 5, 20K, 20K, 25K, Nil})   & 88.7$\pm${\scriptsize 0.3} & {\bf 86.1$\pm${\scriptsize 0.4}} & {\bf 83.1$\pm${\scriptsize 0.6}} & {\bf 81.4$\pm${\scriptsize 0.9}} & {\bf 72.5$\pm${\scriptsize 1.3}} & {\bf 63.5$\pm${\scriptsize 1.9}} \\
\bottomrule
\bottomrule
\end{tabular}
\vspace{-3mm}
\caption{Performance comparison (averaged test accuracy $\pm$ std over 5 runs) of iteratively pruning ResNet-20 on CIFAR-10 dataset using the global weight magnitude pruning method \cite{han2015learning} and the RMSProp optimizer \cite{tieleman2012lecture}.}
\label{per_rms}
\end{table}}

\newpage
\subsection{More Experimental Results on Comparing S-Cyc's $\texttt{max\_lr}$ to well-tuned $\texttt{max\_lr}$}
\label{A4}

we show the performance between S-Cyc's $\texttt{max\_lr}$ to that of an oracle using ResNet-20, global gradient pruning method on CIFAR-10 in Table \ref{lr_table_resnet_20} and ResNet-50, IMP on ImageNet-200 in Table \ref{lr_table_resnet_50}. It can be seen that the $\texttt{max\_lr}$ estimated by S-Cyc falls in the feasible LR region at each pruning cycle, meaning that the validation accuracy of S-Cyc is comparable to the well-tuned $\texttt{max\_lr}$. This highlights the competitiveness of the proposed S-Cyc using ResNet-20 and ResNet-50.

\begin{table}[!ht]
\centering
\setlength\tabcolsep{7.7pt}
\begin{tabular}{lcccccccccc}
\midrule
\midrule
{\small Percent of Weights Remaining, $\lambda$} & 100 & 64 & 51.3 & 41.1 & 32.9 & 21.1 & 5.72\\ \midrule
{\small Well Tuned \texttt{max\_lr} ($\times$\texttt{e}-2)} & 3.4 & 3.2 & 3.8 & 4.6 & 5.6 & 6.2 & 6.8 \\
{\small Feasible LR Region ($\times$\texttt{e}-2)} & {\small [2.8, 3.6]} & {\small [3.0, 3.6]} & {\small [3.4, 4.2]} & {\small [3.8, 5.2]} & {\small [5.4, 6.6]} & {\small [5.4, 6.8]} & {\small [5.8, 7.4]} \\
{\small S-Cyc ($\times$\texttt{e}-2)} & 3 & 3 & 3.2 & 4.7 & 6.4 & 6.9 & 6.99   \\
\bottomrule
\bottomrule
\end{tabular}
\vspace{-3mm}
\caption{Comparison between the value of $\texttt{max\_lr}$ computed by S-Cyc, well-tuned $\texttt{max\_lr}$ via grid search and feasible learning rate region when iteratively pruning the ResNet-20 on the CIFAR-10 dataset using the global weight magnitude pruning method \cite{han2015learning}}
\label{lr_table_resnet_20}
\end{table}

\begin{table}[!ht]
\centering
\setlength\tabcolsep{7.5pt}
\begin{tabular}{lcccccccccc}
\midrule
\midrule
{\small Percent of Weights Remaining, $\lambda$} & 100 & 64 & 51.3 & 41.1 & 32.9 & 21.1 & 5.72\\ \midrule
{\small Well Tuned \texttt{max\_lr} ($\times$\texttt{e}-2)} & 4.8 & 4.6 & 5.4 & 5.6 & 6.8 & 9.4  & 9.2 \\
{\small Feasible LR Region ($\times$\texttt{e}-2)} & {\small [4.6, 5.0]} & {\small [4.6, 5.2]} & {\small [4.6, 5.6]} & {\small [5.2, 6.0]} & {\small [6.6, 7.2]} & {\small [9.0, 10.0]} & {\small [9.0, 10.2]} \\
{\small S-Cyc ($\times$\texttt{e}-2)} & 5 & 5 & 5 & 5.2 & 7.2 & 9.8 & 9.9  \\
\bottomrule
\bottomrule
\end{tabular}
\vspace{-3mm}
\caption{Comparison between the value of $\texttt{max\_lr}$ computed by S-Cyc, well-tuned $\texttt{max\_lr}$ via grid search and feasible learning rate region when iteratively pruning the ResNet-50 on ImageNet-200 dataset using the IMP pruning method \cite{tinyimagenet}.}
\label{lr_table_resnet_50}
\end{table}

\begin{landscape}

\subsection{Experimental Results for More Values of $\lambda$}
\label{A3}
We note that, in Table \ref{per3}, we only show the experimental results for some key values of $\lambda$. In this subsection, we show the results for more values of $\lambda$ in Tables \ref{per1_extra} - \ref{per3_extra}. We observe that, for other values of $\lambda$, the proposed S-Cyc still outperforms selected LR benchmarks.

\begin{table}[!ht]
\centering
\setlength\tabcolsep{21.5pt}
\begin{tabular}{l rrrrr}
\toprule
\toprule
{\small {\bf Params}: 227K} & {\small {\bf Train Steps}: 63K Iters} & {\small {\bf Batch}: 128} & {\small {\bf Batch Norm}: Yes} & {\small {\bf Optimizer}: SGD} & {\small {\bf Rate}: 0.2}\\
\bottomrule
\end{tabular}

\setlength\tabcolsep{4.2pt}
\hspace*{1mm}\begin{tabular}{lccccccccccc}
{\small Percent of Weights Remaining, $\lambda$} & 100 & 80& 64 & 51.3 & 41.1 & 32.9 & 21.1  &  10.9 & 5.72 & 2.03 \\ \midrule
{\small constant LR} ({\footnotesize 1\texttt{e}-2}) &  89.4$\pm${\scriptsize 0.4} & 89.0$\pm${\scriptsize 0.4} & 88.7$\pm${\scriptsize 0.5} & 87.8$\pm${\scriptsize 0.7} & 86.9$\pm${\scriptsize 0.9} & 86.5$\pm${\scriptsize 0.9} & 85.0$\pm${\scriptsize 0.7} & 82.5$\pm${\scriptsize 0.3}  & 78.9$\pm${\scriptsize 0.8} & 66.2$\pm${\scriptsize 1.3}  \\
{\small LR decay} ({\footnotesize 3\texttt{e}-2, 63K}) & 90.0$\pm${\scriptsize 0.4} & 89.3$\pm${\scriptsize 0.2} & 89.2$\pm${\scriptsize 0.3} & 88.0$\pm${\scriptsize 0.5} & 87.2$\pm${\scriptsize 0.6} & 87.0$\pm${\scriptsize 0.8} & 85.7$\pm${\scriptsize 0.6} & 82.9$\pm${\scriptsize 0.6} & 79.6$\pm${\scriptsize 0.7} & 66.9$\pm${\scriptsize 1.9}  \\
{\small cyclical LR} ({\footnotesize 0, 3\texttt{e}-2, 8K}) &  89.6$\pm${\scriptsize 0.3} & 89.2$\pm${\scriptsize 0.3} & 88.7$\pm${\scriptsize 0.5} & 88.2$\pm${\scriptsize 0.4} & 87.7$\pm${\scriptsize 0.6} & 87.2$\pm${\scriptsize 0.6} & 85.6$\pm${\scriptsize 0.5} & 83.0$\pm${\scriptsize 0.4} & 80.1$\pm${\scriptsize 0.8} & 67.8$\pm${\scriptsize 1.4} \\
{\small LR-warmup-1} ({\footnotesize 3\texttt{e}-2, 20K, 20K, 25K, Nil}) & {\bf 90.1$\pm${\scriptsize 0.5}} & 89.6$\pm${\scriptsize 0.3} & 89.4$\pm${\scriptsize 0.3} & 88.2$\pm${\scriptsize 0.5} & 87.8$\pm${\scriptsize 0.7} & 87.4$\pm${\scriptsize 0.8} & 85.5$\pm${\scriptsize 0.6} & 83.4$\pm${\scriptsize 0.6} & 80.0$\pm${\scriptsize 1.4} & 68.9$\pm${\scriptsize 1.3} \\
{\small LR-warmup-2} ({\footnotesize 7\texttt{e}-2, 20K, 20K, 25K, Nil}) & 89.7$\pm${\scriptsize 0.4} & 89.4$\pm${\scriptsize 0.2}  & 89.0$\pm${\scriptsize 0.5} & 89.1$\pm${\scriptsize 0.4} & 88.1$\pm${\scriptsize 0.6}  & 87.6$\pm${\scriptsize 0.5} & 85.9$\pm${\scriptsize 0.4} & 83.8$\pm${\scriptsize 0.8} & 80.6$\pm${\scriptsize 1.2} & 71.3$\pm${\scriptsize 1.2} \\
{\small S-Cyc} ({\footnotesize 3\texttt{e}-2, 4\texttt{e}-2, 1, 5, 20K, 20K, 25K, Nil}) & 90.0$\pm${\scriptsize 0.3} & {\bf 89.6}$\pm${\scriptsize 0.4} & {\bf 89.3$\pm${\scriptsize 0.5}} & {\bf 89.2$\pm${\scriptsize 0.6}} & {\bf 88.3$\pm${\scriptsize 0.4}} & {\bf 88.3$\pm${\scriptsize 0.4}} & {\bf 87.1$\pm${\scriptsize 0.5}} & {\bf 84.5$\pm${\scriptsize 0.8}} & {\bf 81.7$\pm${\scriptsize 0.9}} & {\bf 72.8$\pm${\scriptsize 1.6}} \\
\bottomrule
\bottomrule
\end{tabular}
\label{per_resnet_20}
\vspace{-3mm}
\caption{Performance comparison (averaged test accuracy $\pm$ std over 5 runs) of pruning ResNet-20 on CIFAR-10 dataset using the global gradient pruning method \cite{Blalock2020}. LR-warmup-1 is the standard implementation used in \cite{frankle2018lottery, frankle2020linear}. }
\label{per1_extra}
\end{table}

\begin{table}[!ht]
\centering
\setlength\tabcolsep{21.5pt}
\begin{tabular}{l rrrrr}
\toprule
\toprule
{\small {\bf Params}: 139M} & {\small {\bf Train Steps}: 63K Iters} & {\small {\bf Batch}: 128} & {\small {\bf Batch Norm}: Yes} & {\small {\bf Optimizer}: SGD} & {\small {\bf Rate}: 0.9}\\
\midrule
\end{tabular}

\setlength\tabcolsep{4.25pt}
\hspace*{1mm}\begin{tabular}{lccccccccccc}
{\small Percent of Weights Remaining, $\lambda$} & 100 & 80 & 64 & 51.3 & 41.1 & 32.9 & 21.1  &  10.9 & 5.72 & 2.03 \\ \midrule
{\small constant LR} ({\footnotesize 8\texttt{e}-3}) & 92.0$\pm${\scriptsize 0.3} &  91.1$\pm${\scriptsize 0.4} & 90.5$\pm${\scriptsize 0.5} & 89.0$\pm${\scriptsize 0.6} & 88.3$\pm${\scriptsize 0.6}  & 87.9$\pm${\scriptsize 0.7}& 85.6$\pm${\scriptsize 0.9} & 80.2$\pm${\scriptsize 1.4} & 70.1$\pm${\scriptsize 1.3} & 40.1$\pm${\scriptsize 1.4}  \\
{\small LR decay} ({\footnotesize 1\texttt{e}-2, 63K}) & 92.1$\pm${\scriptsize 0.5} & 91.3$\pm${\scriptsize 0.3}  & 91.1$\pm${\scriptsize 0.5} & 89.2$\pm${\scriptsize 0.5} & 88.7$\pm${\scriptsize 0.4} & 88.4$\pm${\scriptsize 0.5} & 86.1$\pm${\scriptsize 0.6} & 80.3$\pm${\scriptsize 0.8} & 69.8$\pm${\scriptsize 1.9} & 40.9$\pm${\scriptsize 2.9}  \\
{\small cyclical LR} ({\footnotesize 0, 3\texttt{e}-2, 15K})  & 92.3$\pm${\scriptsize 0.6} & 91.6$\pm${\scriptsize 0.4} & 91.3$\pm${\scriptsize 0.6} & 89.6$\pm${\scriptsize 0.4} & 89.2$\pm${\scriptsize 0.5} & 89.2$\pm${\scriptsize 0.4} & 87.2$\pm${\scriptsize 0.8} & 81.1$\pm${\scriptsize 1.3} & 70.4$\pm${\scriptsize 1.6} & 43.2$\pm${\scriptsize 1.9} \\
{\small LR-warmup-1} ({\footnotesize 1\texttt{e}-1, 10K, 32K, 48K, Nil}) & 92.2$\pm${\scriptsize 0.3} & 92.0$\pm${\scriptsize 0.4} & 91.3$\pm${\scriptsize 0.5} & 90.6$\pm${\scriptsize 0.5} & 89.6$\pm${\scriptsize 0.6} & 89.4$\pm${\scriptsize 0.5} & 87.5$\pm${\scriptsize 0.7} & 81.7$\pm${\scriptsize 1.1} & 71.5$\pm${\scriptsize 1.3} & 46.6$\pm${\scriptsize 2.7} \\
{\small LR-warmup-2} ({\footnotesize 4\texttt{e}-2, 10K, 32K, 48K, Nil}) & 92.9$\pm${\scriptsize 0.3} &  {\bf 92.5}$\pm${\scriptsize 0.3} & 92.6$\pm${\scriptsize 0.3}  & 91.2$\pm${\scriptsize 0.7} & 90.1$\pm${\scriptsize 0.4} & 89.5$\pm${\scriptsize 0.4} & 87.0$\pm${\scriptsize 0.5} & 81.4$\pm${\scriptsize 1.2} & 70.9$\pm${\scriptsize 1.4} & 45.0$\pm${\scriptsize 2.3} \\
{\small S-Cyc} ({\footnotesize 4\texttt{e}-2, 6\texttt{e}-2, 2, 4, 10K, 32K, 48K, Nil})  & {\bf 93.0$\pm${\scriptsize 0.4}} & 92.4$\pm${\scriptsize 0.2}  & {\bf 92.8$\pm${\scriptsize 0.4}} & {\bf 91.9$\pm${\scriptsize 0.6}} & {\bf 90.6$\pm${\scriptsize 0.5}} & {\bf 90.2$\pm${\scriptsize 0.7}} & {\bf 88.2$\pm${\scriptsize 1.4}} & {\bf 83.1$\pm${\scriptsize 1.1}} &{\bf 73.2$\pm${\scriptsize 0.9}} & {\bf 48.2$\pm${\scriptsize 2.1}} \\
\bottomrule
\bottomrule
\end{tabular}
\vspace{-3mm}
\caption{Performance comparison (averaged test accuracy $\pm$ std over 5 runs) of pruning VGG-19 on CIFAR-10 dataset using the layer gradient pruning method \cite{Blalock2020}. LR-warmup-1 is the standard implementation used in \cite{frankle2018lottery, frankle2020linear, liu2018rethinking}.}
\label{per2_extra}
\end{table}

\begin{table}[!ht]
\centering
\setlength\tabcolsep{21.3pt}
\begin{tabular}{l rrrrr}
\toprule
\toprule
{\small {\bf Params}: 23.8M} & {\small {\bf Train Steps}: 70K Iters} & {\small {\bf Batch}: 128} & {\small {\bf Batch Norm}: Yes} &  {\small {\bf Optimizer}: SGD} & {\small {\bf Rate}: 0.9} \\
\midrule
\end{tabular}

\setlength\tabcolsep{4.25pt}
\hspace*{1mm}\begin{tabular}{lccccccccccc}
{\small Percent of Weights Remaining, $\lambda$} & 100 & 80 & 64 & 51.3 & 41.1 & 32.9 & 21.1  &  10.9 & 5.72 & 2.03 \\ \midrule
{\small constant LR} ({\footnotesize 1\texttt{e}-2}) & 56.7$\pm${\scriptsize 0.3} & 56.3$\pm${\scriptsize 0.2}  & 55.5$\pm${\scriptsize 0.2} & 55.6$\pm${\scriptsize 0.7} & 55.0$\pm${\scriptsize 0.7}   & 54.4$\pm${\scriptsize 0.4} & 52.7$\pm${\scriptsize 1.2} & 50.1$\pm${\scriptsize 1.3} & 46.3$\pm${\scriptsize 0.9} & 43.9$\pm${\scriptsize 1.2} \\
{\small LR decay} ({\footnotesize 3\texttt{e}-2, 70K} & 56.7$\pm${\scriptsize 0.2} & 56.1$\pm${\scriptsize 0.4} & 56.0$\pm${\scriptsize 0.4} & 55.4$\pm${\scriptsize 0.4}  & 55.2$\pm${\scriptsize 0.5}   & 55.0$\pm${\scriptsize 0.5} & 53.1$\pm${\scriptsize 1.0} & 50.9$\pm${\scriptsize 1.2}  & 47.0$\pm${\scriptsize 0.7} & 41.8$\pm${\scriptsize 1.4}  \\
{\small cyclical LR} ({\footnotesize 0, 5\texttt{e}-2, 20K}) & 56.6$\pm${\scriptsize 0.2} & 56.4$\pm${\scriptsize 0.3} & 55.8$\pm${\scriptsize 0.5} & 55.7$\pm${\scriptsize 0.2}  & 55.8$\pm${\scriptsize 0.3}   & 55.4$\pm${\scriptsize 0.4} & 54.3$\pm${\scriptsize 0.8} & 52.2$\pm${\scriptsize 1.1} & 46.0$\pm${\scriptsize 1.4} & 43.3$\pm${\scriptsize 1.2} \\
{\small LR-warmup-1} ({\footnotesize 1\texttt{e}-1, 4K, 23K, 46K, 62K}) & 57.0$\pm${\scriptsize 0.3} & 57.1$\pm${\scriptsize 0.2} & 56.6$\pm${\scriptsize 0.4} & 56.4$\pm${\scriptsize 0.6} & 56.3$\pm${\scriptsize 0.4} & 56.2$\pm${\scriptsize 0.5} & 56.3$\pm${\scriptsize 0.6} & 53.7$\pm${\scriptsize 0.9} & 49.7$\pm${\scriptsize 1.1} & 47.4$\pm${\scriptsize 1.1} \\
{\small LR-warmup-2} ({\footnotesize 5\texttt{e}-2, 4K, 23K, 46K, 62K}) & {\bf 57.8$\pm${\scriptsize 0.3}} & {\bf 58.1}$\pm${\scriptsize 0.3} & 57.2$\pm${\scriptsize 0.5} & 56.5$\pm${\scriptsize 0.3} & 56.2$\pm${\scriptsize 0.5}  & 56.6$\pm${\scriptsize 0.4} & 55.9$\pm${\scriptsize 0.4} & 53.3$\pm${\scriptsize 1.0} & 48.6$\pm${\scriptsize 1.0} & 45.0$\pm${\scriptsize 0.9} \\
{\small S-Cyc} ({\footnotesize 5\texttt{e}-2, 5\texttt{e}-2, 2, 5, 4K, 23K, 46K, 62K}) &  57.6$\pm${\scriptsize 0.4} & 58.1$\pm${\scriptsize 0.4} & {\bf57.9$\pm${\scriptsize 0.5}} & {\bf57.5$\pm${\scriptsize 0.6}}  & {\bf57.9$\pm${\scriptsize 0.4}}   & {\bf57.1$\pm${\scriptsize 0.5}} & {\bf56.4$\pm${\scriptsize 0.6}} & {\bf54.4$\pm${\scriptsize 1.3}} & {\bf 51.0$\pm${\scriptsize 1.2}} & {\bf 48.9$\pm${\scriptsize 1.5}} \\
\bottomrule
\bottomrule
\end{tabular}
\vspace{-3mm}
\caption{Performance comparison (averaged test accuracy $\pm$ standard deviation over 5 runs) of pruning ResNet-50 on ImageNet-200 dataset using the IMP pruning method \cite{tinyimagenet}. LR-warmup-1 is adapted from the standard implementation used in \cite{frankle2020linear, renda2020comparing}.}
\label{per3_extra}
\end{table}

\end{landscape}

\end{document}